\documentclass[11pt]{article}

\RequirePackage[OT1]{fontenc} \RequirePackage{amsthm,amsmath}
\usepackage{natbib}
\RequirePackage[colorlinks,citecolor=blue,urlcolor=blue]{hyperref}

\usepackage{graphics,epsfig,epsf,psfrag}
\usepackage{url}
\usepackage{bm}
\usepackage{color, appendix}
\usepackage{graphicx}
\usepackage{amsmath, amsfonts, amsthm,amssymb}
\usepackage{epstopdf}
\usepackage{hyperref}
\usepackage{enumerate, framed}
\usepackage{graphicx,lscape,rotate}
\usepackage{algorithmic, algorithm}
\usepackage{multirow}

\let\BLS=\baselinestretch
\makeatletter
\newcommand{\singlespacing}{\let\CS=\@currsize\renewcommand{\baselinestretch}{1}\small\CS}
\newcommand{\doublespacing}{\let\CS=\@currsize\renewcommand{\baselinestretch}{1.5}\small\CS}
\newcommand{\normalspacing}{\let\CS=\@currsize\renewcommand{\baselinestretch}{\BLS}\small\CS}
\makeatother

\numberwithin{equation}{section}
\theoremstyle{plain}
\newtheorem{theorem}{Theorem}[section]

\newtheorem{corollary}{Corollary}[section]
\newtheorem{lemma}{Lemma}[section]
\newtheorem{definition}{Definition}
\theoremstyle{remark}

\def\keywords{\vspace{1.5em}
{\noindent \textbf{Key Words}:\,\,\relax%
}}

\newcommand{\M}{{\mathcal{M}}}

\newcount\colveccount
\newcommand*\colvec[1]{
        \global\colveccount#1
        \begin{pmatrix}
        \colvecnext
}
\def\colvecnext#1{
        #1
        \global\advance\colveccount-1
        \ifnum\colveccount>0
                \\
                \expandafter\colvecnext
        \else
                \end{pmatrix}
        \fi
}

\makeatletter
\newcommand{\Spvek}[2][r]{%
  \gdef\@VORNE{1}
  \left(\hskip-\arraycolsep%
    \begin{array}{#1}\vekSp@lten{#2}\end{array}%
  \hskip-\arraycolsep\right)}

\def\vekSp@lten#1{\xvekSp@lten#1;vekL@stLine;}
\def\vekL@stLine{vekL@stLine}
\def\xvekSp@lten#1;{\def\temp{#1}%
  \ifx\temp\vekL@stLine
  \else
    \ifnum\@VORNE=1\gdef\@VORNE{0}
    \else\@arraycr\fi%
    #1%
    \expandafter\xvekSp@lten
  \fi}
\makeatother

\begin{document}

% \doublespacing % <-- if you want to doublespace your document

\title{Weighted Lasso Estimates for Sparse Logistic Regression: Non-asymptotic Properties with Measurement Error}

\author{\textbf{Huamei Huang}$^1$,~\textbf{Yujing Gao}$^2$,~\textbf{Huiming Zhang}$^3$,~\textbf{Bo Li}$^4$\footnote{ Email: \texttt{haoyoulibo@163.com} (Correspondence author, Prof. Bo Li)},~\textbf{}\footnote{Supported by the National Natural Science Foundation of China (Grant No.61877023) and the Fundamental Research Funds for the Central Universities (grant numbers CCNU19TD009)} \\
\\
\small{1. Department of Statistics and Finance, University of Science and Technology of China, Anhui, 230026, China}\\
\small{2. Guanghua School of Management, Peking University, Beijing, 100871, China}\\
\small{3. School of Mathematical Sciences, Peking University, Beijing, 100871, China}\\
\small{4. School of Mathematics and Statistics, Central China Normal University, Wuhan, 430079, China}}

\date{}  % OR \date{\today}

\maketitle

\renewcommand{\abstractname}{\large Abstract}
\begin{abstract}
When we are interested in high-dimensional system and focus on classification performance, the $\ell_{1}$-penalized logistic regression is becoming important and popular. However, the Lasso estimates could be problematic when penalties of different coefficients are all the same and not related to the data. We proposed two types of weighted Lasso estimates depending on covariates by the McDiarmid inequality. Given sample size $n$ and dimension of covariates $p$, the finite sample behavior of our proposed methods with a diverging number of predictors is illustrated by non-asymptotic oracle inequalities such as $\ell_{1}$-estimation error and squared prediction error of the unknown parameters. We compare the performance of our methods with former weighted estimates on simulated data, then apply these methods to do real data analysis.

\end{abstract}



\section{Introduction}

In recent year, with the advancement of modern science and technology, high-throughput and non-parametric complex data has been frequently collected in gene-biology, chemometrics, neuroscience and other scientific fields. With massive
data in regression problem, we encounter the situation that both the number of covariates $p$ and sample size $n$ are increasing, and $p$ is a function of $n$, i.e. $p=:p(n)$. One further assumption in literatures is that $p$ is allowed to grow with $n$ but $p \le n$, and it has
been extensively studied in subsequent works see \cite{Sur17}, \cite{Fan20},\cite{Zhang2018} and references therein. When we consider the variable selection in terms of linear or generalized linear model, massive data sets bring researchers unprecedented computational challenges, such as the ``large $p$, small $n$'' paradigm, see \cite{Ma2018}. Therefore, another potential characterization appeared in large-scale data is that we only have few significant predictors among $p$ covariates and $p \gg n$. The main challenge is that directly utilizing low-dimensional (classical and traditional) statistical inference and computing methods for these increasing dimension data is prohibitive. Fortunately, the regularized (or penalized) method can perform parameter estimation and variable selection to enhance the prediction accuracy and interpretability of the regression model it generates. One famous proposed method is Lasso (least absolute shrinkage and selection operator), which was introduced in \cite{Tibshirani1996} as modification of least square method in the case of linear models.

In this paper, our tasty for regression covers the case of a binary response (or dichotomous response). The responses $\{ {Y_i}\} _{i = 1}^n$ can take only two values: ``1, 0'',``1, -1'' or some other codes like dichotomous response, such as: good and bad, big and small, win and lose, alive and dead or healthy and sick. This kind of data has been popular for regression in a wide range of applications and fields, including business, computer science (image classification, signal processing), education, and genetic or biomedical fields, see \cite{Guo15}, \cite{Liu2017}, \cite{Li2019} for examples. When it comes to binary regression, logistic regression (logit regression, or logit model) is used every day to perform thousands of classifications, which is by far the most widely used tool for associating binary responses with covariates. Logistic regression has been popular in biomedical research for half a century and it may first be formally studied by statistician David Cox in a discussion paper, see \cite{Cox58}.

Given $n$ covariates $\{X_i\}_{i=1}^{n}\in\mathbb{R}^{p}$, the logistic regression model is of the form
\begin{equation}\label{eq:1}
  P_{\beta^*}(Y_i=1|X_i)=\pi_{\beta^*}(X_i)
  =:\frac{\exp(X_i\beta^*)}{1+\exp(X_i\beta^*)},\quad i=1,2, ..., n,
\end{equation}
where $Y_i\in\{0,1\}$ is the response variable of the individual $i$, $\beta^*$ is a
$p\times1$ vector of unknown regression coefficients belonging to a
compact subset of $\mathbb{R}^p$. The unknown parameter $\beta^*$ is
often estimated by the maximum likelihood estimator (MLE) through
maximizing the log-likelihood function with respect to $\beta$,
namely,
\begin{equation}\label{log-l}
{\hat \beta _{mle}}  = \arg \mathop {\max }\limits_{\beta  \in {\mathbb{R}^p}} \ell (\beta ) = :\arg \mathop {\max }\limits_{\beta  \in {\mathbb{R}^p}} \frac{1}{n}\sum\limits_{i = 1}^n {{\rm{[}}{Y_i}\log \pi_{\beta}(X_i) + (1 - {Y_i})\log \{ 1 - \pi_{\beta}(X_i)\} ]} .
\end{equation}

For more discussions of binary response regression, we refer readers to see \cite{Tutz11} for a comprehensive introduction, and see \cite{Efron2016} which provides a refreshing view of modern statistical inference for today's computer age.

In the high-dimensional case, we often encounter the number of predictors $p$ is larger than the sample size $n$. When $p \gg n$, the least square method leads to over-parameterization, Lasso and many other regularization estimates are required to obtain a stable and satisfactory fitting. Although the Lasso has good performance, some generalized penalities have been proposed since researchers want to compensate for Lasso's certain shortcomings and to make the penalized method more useful for a particular data set, see \cite{Hastie15}. Since Lasso gives the same penalty for each $\beta_j$, an important extension is to use different levels of penalties to shrinkage each covariates' coefficients. The Weighted Lasso estimation method is an improvement of Lasso, where penalized coefficients are estimated based on different data-related weights. But one challenge is that this weighted method depends on bpth covariates and responses, thus it is difficult for us to find the optimal weights.

Many endeavors focus on high-dimensional logistic models (or generalized linear model which includes logistic case) in an attempt to study their weights and sparsity structure in detail. Various weights have been proposed, see for examples: \cite{Geer2008}, \cite{Kwemou16}, \cite{Algamal2017}, \cite{Liu2017}. The contribution of this paper is that we obtain the optimal weight based on the McDiarmid's inequality which is simpler and more feasible than the Bernstein's inequality. The oracle inequality is affected by the measurement error, it is derived under the Stabil Condition with the given level of measurement error. Our work is different from \cite{Kwemou16}, which does not consider conditions with measurement error and their data-related weights contain two unknown constants. The weights proposed in this paper contain only one single unknown constant which is considered as a tuning parameter. Moreover, when $\log p/n$ is not small in \cite{Kwemou16}, the upper bound of $\ell_1$-estimation error oracle inequality is determined by the tuning parameter with the rate $O(\log p/n)$, which is not rate-optimality in the minimax sense. Under some regularized conditions, the optimal convergence rate is $O(\sqrt {\log p/n})$, see chapter 4 of \cite{Rigollet19}. Specifically, our contributions are:
\begin{itemize}

\item This paper proposes the concentration-based weighted Lasso for inference high dimensional sparse logistic regressions, the proposed weights are better than \cite{Kwemou16} in applications.

\item This paper derives the non-asymptotic oracle inequalities with measurement error for weighted Lasso estimates in sparse logistic regressions, and the obtained oracle inequalities are shaper than \cite{Bunea08} in case of Lasso estimates of logistic regressions.

\item The Corollary 2.1 is the theoretical guarantee when we do simulation, the smallest signal should be large than a threshold value.
\end{itemize}

This paper is organized as following. In section 2, the problem of estimating coefficients in logistic regression model by weighted Lasso is discussed, and we give non-asymptotic oracle inequalities for it under the Stabil Condition. In section 3, we introduce a novel data dependent weights for Lasso penalized logistic regression and compare this method with other proposed weights in references. Our weights are based on the event of KKT conditions such that the KKT conditions hold with high probability. In section 4, we use simulation to show our proposed methods can rival the existing weighted estimators, and apply our methods for a real genetic data analysis.

\section{Weighted Lasso Estimates and Oracle Inequalities}

\subsection{$\ell_{1}$-penalized sparse logistic regression}

We consider the following estimator of $\ell_{1}$-penalized logistic regression, which is shown below:

\begin{equation} \label{eq:Wlasso}
\hat{\beta }=\underset{\beta \in \mathbb{R}^p}{\textrm{argmin}}\lbrace - \ell (\beta )+ \lambda \sum_{j=1}^{p} w_j\left| {{\beta _j}} \right|\rbrace =\underset{\beta \in \mathbb{R}^p}{\textrm{argmin}}\left\lbrace - \ell (\beta )+ \lambda || {W\beta } ||_1 \right\rbrace,
\end{equation}
where $\lambda >0$ is the tuning parameter, $\{ {w_j}\} _{j = 1}^p$ are data-dependent weights~(the weights only depend on observed data) and $W = {\rm{diag}}\{ {w_1}, \cdots ,{w_p}\} $ is the $p$-diagonal matrix.

In order to compare with other weighted estimates in reference, we assume these weights are normalized such that  $\sum_{j = 1}^p {{w_j}}  = p$. If all weights $\{w_j\}_{j=1}^p$ in weighted Lasso are chosen as 1, then the estimator degenerates to the ordinary Lasso estimates.

We can also treat $\{ \lambda{w_j}\} _{j = 1}^p$ as a series of tuning parameters. Large value of $\lambda{w_j}$ will bring a smaller estimation of coefficient $\beta_j$, and small value of $\lambda{w_j}$ has the opposite result. We will discuss the chosen of weights in Section 3.

Clearly, a non-sparse estimate of $\beta$ will lead to overfitting, on the contrary, a sparse estimate may capture significant coefficients and eliminate ambiguous variables. Given the training and testing data, the sparse estimate can control the trade-off between the number of response variables and the prediction error of the testing data. Since there is no general closed-form solution to the weighted Lasso estimate
${\hat \beta }$, iterative procedures are often adopted to find numerical solution. The commonly used iterative processes are the Quasi-Newton method and the Coordinate Descent method.

The gradient of $\ell(\beta)$ is
\[\dot \ell (\beta ) = \frac{1}{n}\frac{{\partial \ell (\beta )}}{{\partial \beta }} = \frac{1}{n}\sum\limits_{i = 1}^n {{X_i}} \left\{{Y_i} - \frac{{{\exp\left(X_i\beta \right)}}}{{1  + {\exp\left(X_i\beta\right) }}}\right\} = \frac{1}{n}\sum\limits_{i = 1}^n {{X_i}} \left\{{Y_i} - {\rm{E}}({Y_i}{\rm{|}}{X_i})\right\}.\]

From the KKT conditions (see page 68 of \cite{Buhlmann2011}), the weighted Lasso estimate $\hat{\beta }$ is a solution of (\ref{eq:Wlasso}), and  characterized by the necessary and sufficient conditions showing below:
\begin{eqnarray}\label{eq:kkt}
\left\{
\begin{aligned}
\frac{1}{n}\sum\limits_{i = 1}^n {{X_{ij}}} \left\{{Y_i} - \frac{\exp(X_i\hat \beta )}{{1  + \exp(X_i\hat \beta )}}\right\}= \lambda w_j  \rm{sign}(\hat\beta_{\emph{j}}) \quad\, \text{  if } \hat\beta_{\emph{j}} \neq 0,\\
\left| \frac{1}{n}\sum\limits_{i = 1}^n {{X_{ij}}} \left\{{Y_i} - \frac{{\exp(X_i\hat \beta )}}{{1 + \exp(X_i\hat \beta)}}\right\}\right| \le \lambda w_j  \qquad\quad\text{  if } \hat\beta_{\emph{j}}=0.
\end{aligned}
\right.
\end{eqnarray}

Theoretically, the data-dependent turning parameter $\lambda w_{j}$ is required to ensure the event of KKT conditions evaluated at true parameter holds with high probability. In section 3, we will apply the McDiarmid's inequality of weighted sum of random variables' to obtain the $\lambda w_{j} $. In addition, the McDiarmid's inequality as an important ingredient can help us establish the oracle inequality for weighted Lasso estimate in next section.

\subsection{Oracle inequalities}
\label{asymptotics}

Deriving oracle inequalities is a powerful mathematical skill which provides deep insight into the non-asymptotic fluctuation of an estimator compared to the ideal unknown estimate which is called oracle. A comprehensive theory of high-dimensional regression has been developed for Lasso and its generalization. Chapter 6 of \cite{Buhlmann2011} and Chapter 11 of \cite{Hastie15} outline an overview of the theory of Lasso including works on oracle inequalities. In this section, non-asymptotic oracle inequalities for the weighted Lasso estimate of logistic regression are sought, as well as assumptions of Condition Identif required.

To arrive at our arguments, we first introduce the definition of Condition Identif (a type of restricted eigenvalue originally proposed by \cite{Bickel09}) which provides tuning parameter and sparsity-dependent bounds for the $\ell_{1}$-estimation error and square prediction error. Foremost, we consider the following two assumptions

\begin{itemize}
\item [\textbullet] \textbf{Assumption 1}: Bounded elements of random design matrix. Assume that ${\left\| X \right\|_\infty } =: \mathop {\max }\limits_{i,j} \left| {{X_{ij}}} \right| \le L~ {\rm{a.s.}}$ where $L$ is a positive constant.

\item [\textbullet] \textbf{Assumption 2}: Assume that ${\beta ^*} \in \Lambda  =: \{ \beta  \in {{\mathbb{R}}^p}:||{\beta ^*}|{|_1} \le B\} $.

\end{itemize}

Although there are applications where unbounded covariates will be of interests, for convenience we do not discuss the case with unbounded covariates. We also limit our analysis on bounded predictors since the real data we collected are often bounded. If not bounded we can take a log-transformation of the original data, thus the transformed data is almost bounded. We can also make transformation
 $f(X_{ij})=\exp(X_{ij})/\left\{1+\exp(X_{ij})\right\}$, thus the transformed predictors are undoubtedly bounded variables.

Let ${\beta ^*}$ be the true coefficient, which is defined by minimization of the unknown risk function:
\begin{equation}\label{eq:oracle}
{\beta ^{\rm{*}}}{\rm{ = }}\mathop {{\rm{argmin}}}\limits_{\beta  \in {{\mathbb{R}}^p}} {\rm{E}}\left\{l(Y,X;{{\beta }})\right\}
\end{equation}
where $l(Y,X;{{\beta }})=  - Y{X}\beta  + \log \left\{1 + \exp \left(X\beta\right) \right\}$ is the logistic loss function.

It can be shown that \eqref{eq:oracle} coincides \eqref{eq:1}. The first order condition for convex optimization \eqref{eq:oracle} is
\begin{equation}\label{eq:1rd}
{\rm{0 = }}\frac{\partial }{{\partial \beta }}{\rm{E}}l(Y,X;\beta ){\rm{ = E}}\left\{ {\left[ { - Y + \frac{{\exp \left( {X\beta } \right)}}{{1 + \exp \left( {X\beta } \right)}}} \right]X} \right\}{\rm{ = E}}\left[ {X{\rm{E}}\left\{ {\left. { - Y + \frac{{\exp \left( {X\beta } \right)}}{{1 + \exp \left( {X\beta } \right)}}} \right|X} \right\}} \right],
\end{equation}
thus the population version of \eqref{eq:1} satisfies \eqref{eq:1rd}.

Let $H(\beta^*) = :\{ j:{\beta _j^*} \ne 0\} $ be the non-negative index set, and ${\beta ^*_H} = :\{ {\beta _j^*}:j \in H\} $ be the sparse vector whose index set of non-negative components is $H$. Sometimes, if there is no ambiguity, we write $H(\beta^*)$ as $H$. Define the weighted cone set for any vector $b \in {\mathbb{R}^p}$ as
\begin{equation}\label{eq:re}
{\rm{WC}}(k,\varepsilon ) =: \{b  \in {\mathbb{R}^p}:{\left\| {{W_{{H^{\rm{c}}}}}{b_{{H^{\rm{c}}}}}} \right\|_1} \le k{\left\| {{W_H}{b_H}} \right\|_1} + \varepsilon \},
\end{equation}
which is a weighted and fluctuated (or measurement error) version of the cone condition ${\mathop{\rm S}\nolimits} (s ,H) =: \{  b \in {\mathbb{R}^p}:{|| {{{ b}_{{H^c}}}} ||_1} \le s {|| {{{b}_H}} ||_1}\}$ mentioned in \cite{Bickel09}, where $W_{H}$ is a diagonal matrix with the $j$th diagonal element as $w_{j}$ if $j\in H$, otherwise as $0$.

In the proof, we will put ${b}=\hat\beta- {\beta ^*}$. In real data with measurement error, let $\hat\beta$ be the estimator based on true covariates and $\hat\beta_{me}$ be the estimator from the observed covariates with measurement error (some zero mean random errors). For example, the observed covariates are zero mean normal random variables, but the true covariates are bounded random variables. Under the assumption of the cone condition ${||{{{b}_{{H^c}}}} ||_1} \le c{|| {{{b}_H}}||_1  }$ for ${b}=\hat\beta- {\beta ^*}$, we have
\begin{align*}
|| {{(\hat\beta_{me}- {\beta ^*})_{{H^c}}}}||_1 -||(\hat\beta  -\hat\beta_{me})_{{H^c}}||_1&\le {||{{b_{{H^c}}}} ||_1}\le c{||{{b_{{H}}}} ||_1}\\
 &\le c|| {{(\hat\beta_{me}- {\beta ^*})_H}}||_1+c||(\hat\beta  -\hat\beta_{me})_H||_1.
\end{align*}
Then, we get
\begin{align*}
{||{{{b}^{me}_{{H^c}}}} ||_1} \le c{|| {{{b}^{me}_H}}||_1  } + \varepsilon~~\text{for}~{b}^{me}=:\hat\beta_{me}- {\beta ^*},
\end{align*}
where $\varepsilon=c|| {{(\hat\beta_{me}- {\beta ^*})_H}}||_1+||(\hat\beta  -\hat\beta_{me})_{{H^c}}||_1$. If $\hat\beta_{me}$ is misspecified as $\hat\beta$, this heuristic derivation shows that the fluctuated cone set signifies the level of measurement error.

To establish the desired oracle inequalities, on ${\rm{WC}}(k,\varepsilon )$, we assume that the $p \times p$ matrix $\Sigma={\rm E}(\textit{{X}}\textit{{X}}^{T})$ satisfies at least one of the following conditions: Stabil Condition (see \cite{Bunea08}) and Weighted Stabil Condition (our proposed condition).
\begin{definition} (Stabil Condition) For a given constant $c_2 >0$ and the measurement error $\varepsilon >0$, let $\Sigma={\rm E}(\textit{{X}}\textit{{X}}^{T})$ be a covariance matrix, which satisfies the Stabil condition $S(c_1,\varepsilon,k)$: if there exists $0<k<1$ such that
\[{{b}^T}\Sigma {b} \ge c_1||{{b}_H}||_2^2 - \varepsilon\]
 for any ${b} \in {\rm{WC}}(k,\varepsilon )$.
\end{definition}

\begin{definition} (Weighted Stabil Condition) For a given constant $c_2 >0$ and the measurement error $\varepsilon >0$, let $\Sigma={\rm E}(\textit{{X}}\textit{{X}}^{T})$  be a covariance matrix, which satisfies the Weighted Stabil condition $WS(c_2,\varepsilon,k)$ if there exists $0<k<1$ such that
\[{{b}^T}\Sigma {b} \ge c_2||{{W_H}}{{b}_H}||_2^2 - \varepsilon\]
 for any ${b} \in {\rm{WC}}(k,\varepsilon )$.
\end{definition}
The constants $c_1, c_2$ in the above two conditions are essentially the lower bound on the restricted eigenvalues of the covariance matrix. For convenience, we use the same $\varepsilon$ in the above two conditions and the weighted cone set. Under the above mentioned assumptions, it yields the following oracle inequalities.
\begin{theorem}\label{theo-oracle}
Assume the condition $WS(3,{\varepsilon _n},c_1)$ (or $S(3,{\varepsilon _n},c_1)$), $\rm{Assumption \ 1}$ and $\rm{Assumption\  2}$ are fulfilled. Let ${d^*} = \left| H(\beta^*) \right|$, $A\ge 1$, $s = \frac{{{e^{LB}}}}{{2{{(1 + {e^{LB}})}^2}}}$ and $T(L,B)=: LB + \log (\theta  + {e^{LB}})$. Let ${\lambda}$ be a tuning parameter chosen such that
$\lambda  \ge \frac{{20LA}}{{{w_{\min }}}}\sqrt {\frac{{2\log (2p)}}{n}} .$
Then, with probability at least $1 - {(2p)^{ - {A^2}}}$, we have following results:

1. Fluctuated cone set
$\hat{\beta}-\beta^{*} \in {\rm{WC}}(3,\varepsilon_n/2).$

2. The $\ell_{1}$-estimation error under the $WS(3,{\varepsilon _n},c_1)$ and $S(3,{\varepsilon _n},c_1)$ are
\begin{equation}\label{eq:l1}
||\hat \beta  - {\beta ^*}|{|_1} \le \frac{{2\lambda {d^*}}}{{sk{w_{\min }}}} + \left(\frac{{\lambda  + 2s}}{{\lambda {w_{\min }}}}\right){\varepsilon _n},~~~||\hat \beta  - {\beta ^*}|{|_1} \le \frac{{2\lambda ||{W_H}||_2^2}}{{sk{w_{\min }}}} + \left(\frac{{\lambda  + 2s}}{{\lambda {w_{\min }}}}\right){\varepsilon _n},
\end{equation}
respectively.

3. Suppose that we have a new covariate vector $ X^*$ (as the test data) which is an independent copy of $ X$ (as the training data), and ${\rm{E^*}}$ represents expectation only about $X^*$. If these weights satisfy \begin{equation}\label{eq:star}
B\left(\frac{{4{w_{\max }} + {w_{\min }}}}{{{w_{\min }}}}\right) + \frac{{{\varepsilon _n}}}{w_{\min}} \le {B^*},~{\rm{a.s.}}
\end{equation}
for some constant ${B^*}$, then the square prediction error under the $WS(3,{\varepsilon _n},c_1)$ and $S(3,{\varepsilon _n},c_1)$ are
\begin{equation}\label{eq:l2}
{\rm{E^*}}{\{{ X^{*}}( \hat \beta- {\beta ^*} )\}^2} \le \frac{{3{\lambda ^2}{d^*}}}{{{s^2}k}} + \left(\frac{{2\lambda }}{s} + 3\right){\varepsilon _n},~~~{\rm{E^*}}{\{{ X^{*}}( \hat \beta- {\beta ^*} )\}^2} \le \frac{{3{\lambda ^2}{||{W_H}||_2^2}}}{{{s^2}k}} + \left(\frac{{2\lambda }}{s} + 3\right){\varepsilon _n},
\end{equation}
respectively.
\end{theorem}

\textbf{Remark 1}: If the measurement error is a small order of the optimal rate: ${\varepsilon _n} = o(\sqrt{ \log p/n})$, we have
\begin{equation}
\|\hat \beta  - {\beta ^*}\|_1 \le  O({d^*}\sqrt {\log p /n}) + o(\sqrt{\log p/n}) + o(1)\nonumber
\end{equation}
and
\begin{equation}
{\rm{E^*}}{\{{ X^{*}}( \hat \beta- {\beta ^*} )\}^2} \le O({d^*}\log p/n) + o(\sqrt{\log p/ n}).\nonumber
\end{equation}
 More typical examples for ${\varepsilon _n}$ are $1/n$ or even 0. We observe that when $d^{*}=O(1)$ and the number of covariates increases as large as $o(e^n)$. Then the bound on estimation error is of the order $ o\left(1\right) $ and the weighted Lasso estimator ensures the consistent property. If the measurement error is a big order of the optimal rate: ${\varepsilon _n} > O(\sqrt {\log p/n} )$, then the convergence rate is ${\varepsilon _n}$. If $||{W_H}||_2 \le d^*=\sum_{j = 1}^p {1\{ \beta _j^* \ne 0\} }$, then oracle inequalities under the Weighted Stabil Condition will be more sharp than oracle inequalities based on the Stabil Condition.

\textbf{Remark 2}: If \eqref{eq:Wlasso} is replaced by the robust penalized logistic regression (see \cite{Park2016}, \cite{Yin2020}) as
\begin{equation} \label{eq:WWlasso}
\hat{\beta }=\underset{\beta \in \mathbb{R}^p}{\textrm{argmin}}\left[ \sum\limits_{i = 1}^n {R_i\left[{Y_i}\log {\pi _i}(\beta ) + (1 - {Y_i})\log \{ 1 - {\pi _i}(\beta )\} \right]}+ \lambda \sum_{j=1}^{p} w_j\left| {{\beta _j}} \right|\right],
\end{equation}
where $R_i$ is some weight such that $n{R_i} \le C$ in the weighted log-likelihood (here $C$ is a constant). We still have oracle results similar to Theorem~\ref{theo-oracle}.

\textbf{Remark 3}: Note that $s$-value in Theorem 2.4 of \cite{Bunea08} is $\frac{1}{{{{(1 + {e^{LB}})}^4}}}$. Our $s$-value is $ \frac{{{e^{LB}}}}{{2{{(1 + {e^{LB}})}^2}}}$, which leads to the sharper oracle inequalities \eqref{eq:l1} and \eqref{eq:l2} due to $\frac{{{e^{LB}}}}{{2{{(1 + {e^{LB}})}^2}}} > \frac{1}{{{{(1 + {e^{LB}})}^4}}}$.

The proof of Theorem~\ref{theo-oracle} and the following corollary are both given in Section 5.

Let $\hat H =: \{ j:{{\hat \beta }_j} \ne 0\}$, which is the index set of non-zero components of $\hat \beta$. Now we want to study conditions under which $P(H \subset \hat H) \ge 1 - \delta $ holds for the number of parameters $p$ and confidence $1 - \delta$. By using the Theorem~\ref{theo-oracle}, we are able to bound $P(H \not\subset \hat H)$, then the probability of correct subset selection is $P(H \subset \hat H) \ge 1 - {\delta }$.

\begin{corollary}\label{cl:ci}
Let $\delta  \in (0,1)$ be a fixed number. Suppose that the assumption of Theorem~\ref{theo-oracle} is satisfied, and the weakest signal and strongest signal meet the condition:
\begin{equation}
{{B_0}} =: \frac{{4\lambda {d^*}}}{{2sk}} + \left(\frac{{\lambda  + 2s}}{\lambda }\right){\varepsilon _n} \le \mathop {\min }\limits_{j \in H} |\beta _j^*| \le B\nonumber.
\end{equation}
If $p = \frac{1}{2}\exp\left\{\frac{1}{A^2}\log \left(\frac{1}{\delta}\right)\right\}$, then we get
\[
P(H \subset \hat H) \ge P(\|\hat \beta  - {\beta ^*}\|_1 \le {B_0}) \ge 1 - \delta .
\]
\end{corollary}

This corollary is the theoretical guarantee when we do simulation, the smallest signal of $\beta_j$ should be large that a threshold value which is also called the Beta-min Condition, see \cite{Buhlmann2011}.

\section{Data-dependent Weights}
\label{sec2}

As mentioned before, the weights in equation (\ref{eq:Wlasso}) only depend on the observed data such that KKT conditions hold with high probability. The desired weights result in a weighted Lasso estimates that should have a better convergence rate than the ordinary Lasso estimate. The question is what data-dependent weights can make KKT conditions hold with high probability. The rationale for obtaining data-dependent weights is to properly apply a concentration inequality for weighted sum of independent random variables. And weights can be directly calculated from data, which cannot contain any unknown parameters. For logistic regression, \cite{Kwemou16} designs a criterion to get weights grounded on the Bernstein's concentration inequality, see also \cite{Yang19} for sparse density estimation. However, based on the Bernstein's concentration inequality, the convergence rate of the upper bound is $\exp \left\{-c_1t^2/\left(c_2n+c_3t\right)
\right\}$, which bounds the sum of $n$ independent random variables deviate from its expected value. Compared with the Bernstein's concentration inequality, the McDiarmid's inequality has faster convergence rate $\exp \left(-c_1t^2/n\right)$. The following is the statement of the McDiarmid's inequality, which is also called the bounded difference inequality.

\begin{lemma}\label{lm:bd}
Suppose $X_{1},\cdots,X_{n}$ are independent random variables all taking values in the set $A$, and
assume $f:A^n\rightarrow\mathbb{R}$ is a function satisfying the \emph{bounded difference condition}
$$\underset{x_{1},\cdots,x_{n},x_{k}^{'}\in A}{\textrm{sup}}\vert f(x_{1},\cdots,x_{n})-f(x_{1},\cdots,x_{k-1},x_{k}^{'},x_{k+1},\cdots,x_{n})\vert\le c_{k}.$$
Then for all $t>0$,
\begin{equation}
P\left[ {\left| {f({X_1},\cdots,{X_n}) - {\rm{E}}\left\{f({X_1},\cdots,{X_n})\right\}} \right| \ge t} \right] \le 2\exp\left(
2{t^2}\Big/\sum_{i = 1}^n {c_i^2}\right).\nonumber
\end{equation}
If there are no absolute signs in the above event, then the upper bound is changed by $\exp\left(
2{t^2}/\sum_{i = 1}^n {c_i^2}\right)$.
\end{lemma}
The event of KKT conditions implies
$$\mathcal{K}(\lambda w_j)=: \left\{\left|\frac{1}{n}\sum\limits_{i = 1}^n {{X_{ij}}} \left({Y_i} - \frac{{{e^{{X_i}\beta^* }}}}{\theta  + {{e^{{X_i}\beta^* }}}}\right)\right| \le \lambda {w_j}\right\},\  j = 1,2, \cdots ,p .$$
Now we check the bounded difference condition,
\begin{align*}
&~~~~\frac{1}{n}\left|\sum\limits_{i=1}^n X_{ij}\left(Y_i-\frac{e^{X_i\beta^{*}}}{\theta+e^{X_i\beta^*}}\right)-\left\{\sum\limits_{i=1,i\neq k}^nX_{ij}\left(Y_i-\frac{e^{X_i\beta^*}}{\theta+e^{X_i\beta^*}}\right)+
X_{kj}\left(Y_k-\frac{e^{X_k\beta^*}}{\theta+e^{X_k\beta^*}}\right)
\right\}
\right|\\
& \le \frac{1}{n}\left| {{X_{ij}}\left({Y_i} - \frac{{{e^{{X_i}\beta^* }}}}{{\theta  + {e^{{X_i}\beta^* }}}}\right) - {X_{kj}}\left({Y_k} - \frac{{{e^{{X_k}\beta^* }}}}{{\theta  + {e^{{X_k}\beta^* }}}}\right)} \right|\le \frac{1}{n}\left( {\left| {{X_{ij}}} \right|{\rm{ + }}\left| {{X_{kj}}} \right|} \right) \le \frac{2}{n}\mathop {\max }\limits_{k = 1, \cdots ,n} \left| {{X_{kj}}} \right|.
\end{align*}

The McDiarmid's inequality gives
\begin{equation}\label{eq:McDiarmid}
P\left\{ {\left| {\frac{1}{n}\sum\limits_{i = 1}^n {{X_{ij}}} \left({Y_i} - \frac{{{e^{{X_i}\beta^* }}}}{{\theta  + {e^{{X_i}\beta^* }}}}\right)} \right| \ge \lambda {w_j}} \right\} \le 2\exp \left\{  - \frac{{n{{(\lambda {w_j})}^2}}}{{2\mathop {\max }\limits_{k = 1, \cdots ,n} {{\left| {{X_{kj}}} \right|}^2}}}\right\}  = :{p^{ - r}},
\end{equation}
where $r> 0$ is a constant.

Therefore, we get
\begin{equation}\label{eq:weights}
\lambda {w_j} = \mathop {\max }\limits_{k = 1, \cdots ,n} \left| {{X_{kj}}} \right|\sqrt {\frac{2}{n}(r\log p + \log 2)}.
\end{equation}

We set data-dependent weights as $w_j,\ j = 1,2, \cdots ,p$ from equation $(3.2)$. It should be noted that these weights obtained in \cite{Kwemou16} are
\begin{equation}
\omega_{j}=\frac{2}{n} \sqrt{\frac{1}{2} \sum_{i=1}^{n} \phi_{j}^{2}\left(z_{i}\right)(x+\log p)}+\frac{2 c(x+\log p)}{3 n},~(x,c \ge 0\nonumber, \ j=1,2,\cdots,p.)
\end{equation}
which may not obtain the optimal tuning parameter $\lambda {w_j} = O(\sqrt {\log p/n})$ when $\log p/n$ is large (as the main term), and the constant $c$ is usually unknown in the real data (the $c$ should be treated as the second tuning parameter, so it results in a large amount of computation).

One may also find more particular way to define weights in references. In high-dimensional settings, unlike the previous proposed weighted Lasso estimates, our weights based on conditions which can be checked to hold with high probability. Various weights that we compare in next section are listed below:
\begin{itemize}
\item[-] Type I Weight (Proposed concentration based weight I):
$$w_j^{(1)} \propto  \mathop {\max }\limits_{k = 1, \cdots ,n} \left| {{X_{kj}}} \right|\sqrt {\frac{2}{n}(r\log p + \log 2)},~~r = 1;$$
\item[-] Type II Weight  (Proposed concentration based weight II):
$$
w_j^{(2)} \propto \sqrt {\frac{1}{n}\sum\limits_{k = 1}^n {X_{kj}^2} }  \cdot \sqrt {\frac{2}{n}(r\log p + \log 2)},~~r = 1;$$
\item[-] Type III Weight  (Inverse standard deviation weights, \cite{Algamal2017}): $
w_j^{(3)} \propto \left\{{{\frac{1}{n}\sum\limits_{i = 1}^n {{{\left( {{X_{ij}} - {{\bar X}_{ \cdot j}}} \right)}^2}} } }\right\}^{-1};$
\item[-] Type IV Weight  (Adaptive Lasso, \cite{Zou2006}): $
w_j^{(4)} \propto | {\hat \beta _j^{lasso}}|^{-1}.$
\end{itemize}
\section{Simulation and Real Data Results}
\subsection{Simulation Results}

In this section, we compare the performance of the ordinary Lasso estimate and the weighted Lasso estimate of logistic regression on simulated data sets. We use \texttt{R} package \texttt{glmnet} with function \texttt{glmreg()} to fit the ordinary Lasso estimate of logistic regression. For the weighted Lasso estimate, we first apply function \texttt{cv.glmnet()} for 10-fold cross-validation to obtain the optimal tuning parameter $\lambda_{op}$. The actual weights we use are the standardized weights given by
\begin{equation}
\tilde w_j =: \frac{{pw_j}}{{\sum\nolimits_{j = 1}^p {w_j} }}.\nonumber
\end{equation}

Then we transfer our weighted problem into unweighted problem and apply the function \texttt{lbfgs()} in \texttt{R} package \texttt{lbfgs} to find the solution for the unweighted Lasso optimal problem. The original weighted optimal problem is shown below
\begin{equation} \label{eq:Wlasso1}
\hat{\beta }^{(t)}=\underset{\beta \in \mathbb{R}^p}{\textrm{argmin}}\{ - \ell (\beta )+ {\lambda _{{\rm{op}}}} \sum_{j=1}^{p} \tilde w_j\left| {{\beta _j}} \right|\}
\end{equation}

We replace the observation matrix $X$ by the transformed matrix $\tilde{X}$, where $$\tilde{X}_{\cdot j}=:X_{\cdot j}/w_{j},\ j=1,2,\cdots,p.$$

Then we apply function \texttt{lbfgs()} to the transformed matrix $\tilde{X}$ and original outcome $Y$ to get the unweighted Lasso estimate $\tilde{\beta}$. Finally, we transfer $\tilde{\beta}$ into $\hat{\beta}$ by
$$\hat{\beta}_j=:\tilde{\beta}_j/w_j,\ j=1,2,\cdots,p.$$

We simulate 100 random data sets as training data and 200 random data sets as testing data. Training data and testing data are generated as shown in the next section. By optimization with suitable $\lambda_{op}$, we obtain the model with the parameter  $\hat\beta_{\lambda_{\textrm{op}}}$. We calculate the performance of $\ell_1$ estimation error $\Vert \beta^{*} - \hat\beta_{\lambda_{\textrm{op}}} \Vert_{1}$ and prediction error on the testing data sets ($X_{\textrm{test}}$ of size $n_{\textrm{test}}$) by the average of the 100-times errors, where the prediction error is defined as
\begin{align*}
\sqrt{\frac{1}{100}\times\Vert X_{\textrm{test}} \beta^{*} - X_{\textrm{test}} \hat\beta_{\lambda_{\textrm{op}}} \Vert_{2}}
\end{align*}
\\
\textbf{A. Data generation}\\

For each simulation, we set $n_{\textrm{train}} = 100,\  n_{\textrm{test}} = 200$. We set dimension as $p = 50,\ 100,\ 150,\ 200$, and adopt the simulation setting as these following two patterns.
\begin{enumerate}
\item The predictor variables $X$ are randomly drawn from the multivariate normal distribution $\mathcal{N}_{p}(0, \Sigma)$, where $\Sigma$ has elements $\rho^{|k - l|}~(k, l = 1, 2, \cdots, p)$. The correlation among predictor variables are controlled by $\rho$ with $\rho =0.3,\  0.5$ and 0.8. We assign the true coefficient parameter of logistic regression as
\begin{equation*}
\beta^* =(\underbrace{10, \cdots, 10}_{9}, \underbrace{0, \cdots, 0}_{p - 9} ).
\end{equation*}
\item Similar to case 1, we generate predictor variables $X$ from multivariate normal distribution $\mathcal{N}_{p}(0, \Sigma)$, where $\Sigma$ has elements $\rho^{|k - l|}~(k, l = 1, 2, \cdots, p)$, and $\rho =0.3,\  0.5$ and $0.8$, and set the true coefficient parameter as
\begin{equation*}
\beta^*= ( \underbrace{17, \cdots, 17}_{3}, \underbrace{-5, \cdots, -5}_{3}, \underbrace{7, \cdots, 7}_{3}, \underbrace{0, \cdots, 0}_{p - 9} ).
\end{equation*}
\end{enumerate}
\textbf{B. Simulation results}\\

These simulation results are listed in Table ~\ref{table1} to Table ~\ref{table3}.
\begin{table}\footnotesize
	\centering
	\caption{Means of $\ell_1$-error and predition error for simulation 1 and 2 ($\rho=0.3$)}
	\label{table1}
	\begin{tabular}{lcc|lcc}
		\hline
		Method&$||\hat{\beta}-\beta^*||_1$&Prediction error&	Method&$||\hat{\beta}-\beta^*||_1$&Prediction error\\
		\hline
		\multicolumn{3}{c|}{p=50, simulation 1}&	\multicolumn{3}{c}{p=50, simulation 2}\\
		Lasso&1.63365&34.28534 &Lasso&1.93251&38.70548\\
		Type I Weight&1.42433&24.23225&Type I Weight&1.70240&28.53307\\
		Type II Weight&1.41866&24.16879&Type II Weight&1.69252&28.42235\\
		Type III Weight&1.42011&24.19330&Type III Weight&1.69310&28.43208\\
		Type IV Weight&3.53346&68.98107&Type IV Weight&2.01674&36.98280\\
		\multicolumn{3}{c|}{p=100, simulation 1}&	\multicolumn{3}{c}{p=100, simulation 2}\\
		Lasso&0.84767&35.39955&Lasso&1.00386&40.75697\\
		Type I Weight&0.82936&28.83360&Type I Weight&0.98829&33.81417\\
		Type II Weight&0.82611&28.76972&Type II Weight&0.98646&33.78084\\
		Type III Weight&0.82623&28.77004&Type III Weight&0.98692&33.79462\\
		Type IV Weight&2.17210&82.87875&Type IV Weight&1.41508&52.05037\\
		\multicolumn{3}{c|}{p=150, simulation 1}&	\multicolumn{3}{c}{p=150, simulation 2}\\
		Lasso&0.57387&35.97669&Lasso&0.67668&40.45453\\
		Type I Weight&0.58118&30.61218&Type I Weight&0.68632&34.87408\\
		Type II Weight&0.57954&30.56721&Type II Weight&0.68436&34.81442\\
		Type III Weight&0.57991&30.57543&Type III Weight&0.68456&34.81883\\
		Type IV Weight&1.81343&98.65162&Type IV Weight&0.94153&47.99915\\
		\multicolumn{3}{c|}{p=200, simulation 1}&	\multicolumn{3}{c}{p=200, simulation 2}\\
		Lasso&0.43394&36.43936&Lasso&0.51223&41.17015\\
		Type I Weight&0.44563&31.58401&Type I Weight&0.52906&36.22489\\
		Type II Weight&0.44476&31.56000&Type II Weight&0.52788&36.18576\\
		Type III Weight&0.44480&31.56915&Type III Weight&0.52861&36.21349\\
		Type IV Weight&1.30047&92.25202&Type IV Weight&0.84236&57.18696\\
		\hline
	\end{tabular}
\end{table}

\begin{table}\footnotesize
	\centering
	\caption{Means of $\ell_1$-error and predition error for simulation 1 and 2 ($\rho=0.5$)}
	\label{table2}
	\begin{tabular}{lcc|lcc}
		\hline
		Method&$||\hat{\beta}-\beta^*||_1$&Prediction error&	Method&$||\hat{\beta}-\beta^*||_1$&Prediction error\\
		\hline
		\multicolumn{3}{c|}{p=50, simulation 1}&	\multicolumn{3}{c}{p=50, simulation 2}\\
		Lasso&1.62804&41.71149&Lasso&1.94525&42.19085\\
		Type I Weight&1.43439&30.26700&Type I Weight&1.74682&31.01396\\
		Type II Weight&1.43009&30.23423&Type II Weight&1.74024&30.92691\\
		Type III Weight&1.43155&30.24629&Type III Weight&1.74254&30.95419\\
		Type IV Weight&3.96514&94.80968&Type IV Weight&1.73099&32.88244\\
		\multicolumn{3}{c|}{p=100, simulation 1}&	\multicolumn{3}{c}{p=100, simulation 2}\\
		Lasso&0.84851&43.50948&Lasso&1.00330&43.79428\\
		Type I Weight&0.82767&35.76218&Type I Weight&0.98572&36.04177\\
		Type II Weight&0.82409&35.68422&Type II Weight&0.98351&35.99504\\
		Type III Weight&0.82445&35.69734&Type III Weight&0.98370&36.00381\\
		Type IV Weight&2.24129&101.81468&Type IV Weight&1.21341&47.26953\\
		\multicolumn{3}{c|}{p=150, simulation 1}&	\multicolumn{3}{c}{p=150, simulation 2}\\
		Lasso&0.57206&44.15909&Lasso&0.67645&44.08771\\
		Type I Weight&0.57178&37.48219&Type I Weight&0.68535&37.82445\\
		Type II Weight&0.57024&37.43615&Type II Weight&0.68359&37.77004\\
		Type III Weight&0.57033&37.44478&Type III Weight&0.68388&37.77792\\
		Type IV Weight&1.62907&108.31138&Type IV Weight&0.98612&58.47258\\
		\multicolumn{3}{c|}{p=200, simulation 1}&	\multicolumn{3}{c}{p=200, simulation 2}\\
		Lasso&0.43170&44.06943&Lasso&0.51023&44.64566\\
		Type I Weight&0.43662&38.06866&Type I Weight&0.52472&38.98144\\
		Type II Weight&0.43597&38.05382&Type II Weight&0.52373&38.93973\\
		Type III Weight&0.43621&38.06604&Type III Weight&0.52379&38.94242\\
		Type IV Weight&1.23980&107.16770&Type IV Weight&0.86637&67.34089\\
		\hline
	\end{tabular}
\end{table}

From these simulation results, our proposed estimate with Type II weight is better than other methods in most cases, both in $\ell_1$-error and prediction error. The adaptive Lasso estimate performs worst among these five $\ell_1$-penalized estimates.

\begin{table}\footnotesize
	\centering
	\caption{Means of $\ell_1$-error and predition error for simulation 1 and 2 ($\rho=0.8$)}
	\label{table3}
	\begin{tabular}{lcc|lcc}
		\hline
		Method&$||\hat{\beta}-\beta^*||_1$&Prediction error&	Method&$||\hat{\beta}-\beta^*||_1$&Prediction error\\
		\hline
		\multicolumn{3}{c|}{p=50, simulation 1}&	\multicolumn{3}{c}{p=50, simulation 2}\\
		Lasso&1.63073&59.36516&Lasso&1.94580&48.91084\\
		Type I Weight&1.45328&43.50855&Type I Weight&1.77261&34.66073\\
		Type II Weight&1.44921&43.44339&Type II Weight&1.76981&34.61648\\
		Type III Weight&1.45401&43.51757&Type III Weight&1.77388&34.68965\\
		Type IV Weight&4.04162&121.25215&Type IV Weight&3.34649&86.41294\\
		\multicolumn{3}{c|}{p=100, simulation 1}&	\multicolumn{3}{c}{p=100, simulation 2}\\
		Lasso&0.83989&60.82986&Lasso&1.00186&51.65554\\
		Type I Weight&0.79627&49.25652&Type I Weight&0.98286&40.76257\\
		Type II Weight&0.79476&49.23149&Type II Weight&0.98004&40.73804\\
		Type III Weight&0.79522&49.26275&Type III Weight&0.98025&40.74769\\
		Type IV Weight&2.14136&126.74045&Type IV Weight&1.76386&88.57558\\
		\multicolumn{3}{c|}{p=150, simulation 1}&	\multicolumn{3}{c}{p=150, simulation 2}\\
		Lasso&0.56722&63.01125&Lasso&0.67311&52.29483\\
		Type I Weight&0.55227&52.83520&Type I Weight&0.67518&42.95876\\
		Type II Weight&0.55078&52.77857&Type II Weight&0.67393&42.92367\\
		Type III Weight&0.55121&52.80501&Type III Weight&0.67435&42.94123\\
		Type IV Weight&1.51633&130.63107&Type IV Weight&1.31221&103.95876\\
		\multicolumn{3}{c|}{p=200, simulation 1}&	\multicolumn{3}{c}{p=200, simulation 2}\\
		Lasso&0.42844&62.92070&Lasso&0.50784&52.64225\\
		Type I Weight&0.42486&54.22954&Type I Weight&0.51764&44.25601\\
		Type II Weight&0.42367&54.16917&Type II Weight&0.51661&44.21604\\
		Type III Weight&0.42387&54.18049&Type III Weight&0.51694&44.24049\\
		Type IV Weight&1.22850&134.19107&Type IV Weight&1.01761&107.86477\\
		\hline
	\end{tabular}
\end{table}

\subsection{Real Data Results}
In this section, we apply our proposed estimates to analyze biological data. We consider the following two complete data sets.\\
(a) The first data set is the gene expression data from a leukemia microarray study (see \cite{Golub1999}). The data comprises $n=72$ patients, in which 25 patients have acute myeloid leukemia (AML) and 47 patients have acute lymphoblastic leukemia (ALL). Therefore, the binary response in this data set is categorized as AML (label 0) and ALL (label 1). The predictors are the expression levels of $p=7129$ genes. The data can be found in R package \texttt{golubEsets}.\\
(b) The second data set is the above gene expression data after preprocessing and filtering (see \cite{Dudoit2002}). This process reduces the number of genes to $p=3571$. The data set can be found in R package \texttt{cancerclass}.\\

Our target is to select useful genes for specifying AML and ALL. Note that there is no available information about model parameters, we cannot directly compare the selection accuracy and prediction accuracy.  Therefore, we list model size and prediction error of estimated model under Leave-One-Out Cross-Validation (LOOCV) framework. Model size can show the coverage of current estimated model, prediction error can show the prediction accuracy. We apply the ordinary Lasso method and four different weighted Lasso methods as described in the previous section to analyze these data sets. Since lots of coefficients estimated by weighted Lasso methods are small but not zero, we choose $10^{-4}$ and $10^{-5}$ as the limits for these two data sets separately, and set the coefficients less than these limits to zero. These results are summarized in Table ~\ref{table4}.
\begin{table}\footnotesize
	\centering
	\caption{Mean and standard deviation of model size and misclassification rate under Leave-One-Out Cross-Validation framework}
	\label{table4}
	\begin{tabular}{lcc}
		\hline
	   Method &Model size& misclassification rates\\
		\hline
		 &\multicolumn{2}{c}{(a) Gene expression data with $p=7129$ genes, limit=$10^{-4}$}\\
		 \cline{2-3}
		Lasso&24.58(2.78)&0.47(0.50)\\
		Type I Weight&73.17(37.09)&0.32(0.47)\\
		Type II Weight &24.78(7.60)&0.32(0.47)\\
		Type III Weight &2.61(1.41)&0.47(0.50)\\
		Type IV Weight &9.22(2.61)&0.35(0.48)\\
		\hline
		&\multicolumn{2}{c}{(a) Gene expression data with $p=7129$ genes, limit=$10^{-5}$}\\
		\cline{2-3}
		Lasso&24.83(2.72)&0.49(0.50)\\
		Type I Weight&474.50(288.08)&0.35(0.48)\\
		Type II Weight&326.29(252.62)&0.35(0.48)\\
		Type III Weight&12.32(5.05)&0.31(0.46)\\
		Type IV Weight&40.75(19.66)&0.35(0.48)\\
		\hline
		 &\multicolumn{2}{c}{(b) Gene expression data with $p=3571$ genes, limit=$10^{-4}$}\\
		 \cline{2-3}
		Lasso&25.13(2.77)&0.43(0.50)\\
		Type I Weight&32.40(10.74)&0.35(0.48)\\
		Type II Weight&24.68(7.56)&0.35(0.48)\\
		Type III Weight&0.82(0.68)&0.49(0.50)\\
		Type IV Weight&11.96(0.50)&0.35(0.48)\\
		\hline
		&\multicolumn{2}{c}{(b) Gene expression data with $p=3571$ genes, limit=$10^{-5}$}\\
		\cline{2-3}
		Lasso&25.13(2.54)&0.46(0.50)\\
		Type I Weight&216.75(112.57)&0.35(0.48)\\
		Type II Weight&192.22(90.67)&0.35(0.48)\\
		Type III Weight&10.36(4.21)&0.34(0.47)\\
		Type IV Weight&42.85(18.57)&0.35(0.48)\\
		\hline
	\end{tabular}
\end{table}
 In the first data set with limit $10^{-4}$, weighted Lasso estimate with Type II Weight has the best prediction performance; in the first data set with limit $10^{-5}$, weighted Lasso estimate with weight Type III Weight has the best prediction performance. In the second data set with limit $10^{-4}$, weighted Lasso estimates with Type I Weight, Type II Weight and Type IV Weight have similar prediction performance, and weighted Lasso estimates with Type II Weight and Type IV Weight use less predictor variables than other methods; in the second data set with limit $10^{-5}$, weighted Lasso estimate with Type III Weight has the best prediction performance also use the fewest variables. Therefore, we can see that weighted Lasso method with Type II Weight, Type III Weight and Type IV Weight can estimate more accurately with less predictors than other methods.\\

After comparing these five methods, we  build model for the complete observations, and report the selected genes in ordinary Lasso regression and some weighted Lasso methods which have small LOOCV errors. These results are listed in Table ~\ref{table5} and Table ~\ref{table6}. We observe that weighted Lasso estimate with Type II Weight selects more variables than weighted Lasso estimates with Type III Weight and Type IV Weight, and has less refitting prediction error. Summarize these above results, our proposed weighted Lasso method can pick much more meaningful variables for explaining and prediction.

\begin{table}\footnotesize
	\newcommand{\tabincell}[2]{\begin{tabular}{@{}#1@{}}#2\end{tabular}}%
	\centering
	\caption{(a) Genes associated with ALL and AML selected by four methods ($p=7129$)}
	\label{table5}
	\begin{tabular}{c|c|c|c|c}
		\hline
		&Lasso&Type II Weight($10^{-4}$)&Type III Weight($10^{-5}$)& Type IV Weight($10^{-5}$)\\
		\hline
	    Model size&25&29&12&8\\
	    \hline
	     \tabincell{c}{Refitting \\ Prediction Error}&0.54&0.00&0.21&0.00\\
	    \hline
	    \multirow{29}{*}{Selected genes}
	    & HG1612-HT1612\_at&D50913\_at&L13278\_at&HG1612-HT1612\_at\\
	    &L07633\_at&D88270\_at&U43885\_at&J04164\_at
\\
	    &M11147\_at&HG1612-HT1612\_at&U60205\_at&M11722\_at
\\
	    &M16038\_at&J04101\_at&U77948\_at&M17733\_at
\\
	    &M19507\_at&J04164\_at&U89942\_at&M26602\_at
\\
	    &M20902\_at&M11722\_at&X63469\_at&M92287\_at
\\
	    &M23197\_at&M12759\_at&Y08612\_at&U05259\_rna1\_at
\\
	    &M27891\_at&M21624\_at&Y10207\_at&X57351\_s\_at
\\
	    &M31303\_rna1\_at&M26602\_at&M61853\_at&\\
	    &M31994\_at&M38690\_at&M96843\_at&\\
	    &M63138\_at&M89957\_at&HG2562-HT2658\_s\_at	&\\
	    &M84526\_at&M92287\_at&M84371\_rna1\_s\_at&\\
	    &U82759\_at&S76617\_at	&&\\
	    &X17042\_at&U05259\_rna1\_at&&\\		
	    &X59417\_at&U33822\_at&&\\
	    &X95735\_at&U65932\_at&&
\\
	    &Y07604\_at&X51521\_at	&&\\
	    &Y08612\_at&X59417\_at	&&
\\
	    &Y10207\_at&X69111\_at	&&
\\
	    &M21535\_at&X82240\_rna1\_at&&\\	
	    &HG2562-HT2658\_s\_at&X99920\_at&&
\\
	    &M13690\_s\_at&Y10207\_at&&\\
	    &M84371\_rna1\_s\_at&X58072\_at&&\\	
	    &X85116\_rna1\_s\_at&D00749\_s\_at&&\\	
	    &M31523\_at&L06797\_s\_at&&
\\
	    &&U89922\_s\_at&&\\
	    &&U37055\_rna1\_s\_at&&\\		
	    &&M12959\_s\_at&
&\\
	    &&M84371\_rna1\_s\_at&&\\
	    \hline
	\end{tabular}
\end{table}
\begin{table}\footnotesize
	\newcommand{\tabincell}[2]{\begin{tabular}{@{}#1@{}}#2\end{tabular}}%
		\centering
		\caption{(b) Genes associated with ALL and AML selected by four methods ($p=3571$)}
		\label{table6}
		\begin{tabular}{c|c|c|c|c}
			\hline
			&Lasso&Type II Weight($10^{-4}$)&Type III Weight($10^{-5}$)& Type IV Weight($10^{-4}$)\\
			\hline
			Model size&26&30&12&12\\
			\hline
			 \tabincell{c}{Refitting \\ Prediction Error}&0.49&0.00&0.64&0.00\\
			\hline
			\multirow{31}{*}{Selected genes}&HG1612-HT1612\_at
		&D50913\_at&HG2855-HT2995\_at&\tabincell{c}{AFFX-HUMGAPDH/\\M33197\_M\_at} \\
			&L07633\_at&D88270\_at&M17754\_at&J04164\_at\\
			&M11147\_at&HG1612-HT1612\_at&M31303\_rna1\_at&M11722\_at
\\
			&M16038\_at&HG4662-HT5075\_at&U09860\_at&M16279\_at
\\
			&M19507\_at&J04101\_at&U76272\_at&M26602\_at
\\
			&M20902\_at&J04164\_at&U77948\_at&M33680\_at
\\
			&M23197\_at&M11722\_at&X63469\_at&U05259\_rna1\_at\\
			&M27891\_at&M12759\_at&X63753\_at&U51240\_at
\\
			&M31303\_rna1\_at&M21624\_at&Y10207\_at&Z23090\_at\\
			&M31627\_at&M26602\_at&M96843\_at&L06797\_s\_at
\\
			&M31994\_at&M38690\_at&M84371\_rna1\_s\_at&M14483\_rna1\_s\_at
\\
			&M63138\_at&M89957\_at&M65214\_s\_at&U06155\_s\_at
\\
			&M84526\_at&M92287\_at
&&\\
			&U50136\_rna1\_at&S76617\_at&&\\	
			&U82759\_at&U02031\_at	&&\\
			&X17042\_at&U05259\_rna1\_at&&\\		
			&X59417\_at&U33822\_at	&&\\	
			&X63469\_at&U65932\_at	
&&\\
			&X95735\_at&X03934\_at	
&&\\
			&Y07604\_at&X51521\_at	
&&\\
			&Y08612\_at&X59350\_at	
&&\\
			&Y10207\_at&X69111\_at	
&&\\
			&M13690\_s\_at&X99920\_at	&&
\\
			&M84371\_rna1\_s\_at&Y10207\_at	&&
\\
			&X85116\_rna1\_s\_at&Z70220\_at&&	
\\
			&M31523\_at&M96843\_at&&\\
			&&X58072\_at&&
\\
			&&U89922\_s\_at	&&\\
			&&M84371\_rna1\_s\_at&&
\\
			&&M97796\_s\_at&&\\
			\hline
		\end{tabular}
	\end{table}

\section{Proofs}
\subsection{The Proof of Theorem~\ref{theo-oracle}}

Non-asymptotic analysis of Lasso and its generalization often lean on several steps:

First, propose an restricted eigenvalue condition or other analogous condition about the design matrix which guarantee a form of local orthogonality via a restricted set of coefficient vector.

The second step is to get the size of tuning parameter based on KKT optimality conditions (or other KKT-like condition such as Dantzig selector); third, according to restricted eigenvalue assumptions and tuning parameter selection, derive the oracle inequalities via the definition of Lasso optimality and the minimizer under unknown expected risk function and some basic inequalities.

The third step is further divided into 3 sub-steps: (i) Under the KKT-like conditions, show that $\hat \beta- \beta^{*}$ is in some restricted set, and check $\hat \beta- \beta^{*}$ is in a big compact set; (ii) Show that the likelihood-based divergence of $\hat \beta$ and $\beta^{*}$ can be lower bounded by some quadratic distance between $\hat \beta$ and $\beta^{*}$; (iii) By some elementary inequalities and (ii), show that the $||\hat \beta- \beta^{*}||_1$ is in a smaller compact set with radius of optimal rate (proportional to the tuning parameter).

 Our language of proof is heavily impacted by theory of empirical process. For simplicity, we denote the theoretical risk by $\mathbb{P}l({{\beta }}) =:{\rm{E}}\left\{l(y,\beta ,X)\right\}$ and the empirical risk by
$$\mathbb{P}_{n}l({{\beta }})=:- \frac{1}{n}\sum\limits_{i = 1}^n \left[Y_iX_i\beta  - \log \left\{1 + \exp\left(X_i\beta \right)\right\} \right].$$

Merely using the definition of weighted Lasso estimates $\hat{\beta}$ (see equation \eqref{eq:Wlasso}), we get
\begin{equation}\label{eq:def}
{\mathbb{P}_n}l(\hat \beta ) + {\lambda }||W\hat \beta||_1 \le {\mathbb{P}_n}l(\beta^{*}) + \lambda ||W{\beta ^*} ||_1.
\end{equation}

By adding $\mathbb{P}\{l(\hat \beta )- l(\beta^{*})\}+\frac{\lambda}{2} ||W(\hat \beta  - {\beta ^*})||_1$ to both sides of equation (\ref{eq:def}) and rearrange the new inequality, we have
\begin{equation}\label{eq:def-putting}
\mathbb{P}\{l(\hat \beta )- l(\beta^{*})\}+ \frac{\lambda}{2}  ||W(\hat \beta  - {\beta ^*})||_1 \le ({\mathbb{P}_n} - \mathbb{P})\{l({\beta ^*})-l(\hat \beta )\} +\frac{\lambda}{2}  ||W(\hat \beta  - {\beta ^*})||_1+ \lambda (||W{\beta ^*} ||_1 - ||W\hat \beta|{|_1}).
\end{equation}

The following proof is divided by three steps.

\subsection{Step1: Choosing the order of tuning parameter}
Define the following stochastic Lipschitz constant in terms of the suprema of centralized empirical process
\begin{equation}
Z_n({\beta ^*}) =:\underset{\beta\in \mathbb{R}^p}{\sup} \frac{{({\mathbb{P}_n} - \mathbb{P})\left\{ {{l}({\beta ^*}) - {l}(\beta)} \right\}}}{{ ||W( \beta  - {\beta ^*})|{|_1}  + {\varepsilon }}}, \nonumber
\end{equation}
which is a random function of $\{ ({Y_i},{X_i})\} _{i = 1}^n$.

Next, for applying the McDiarmid's inequality to $Z_n({\beta ^*})$, we will show the bounded difference condition if we replace the $i$-th pair $({Y_j},{X_j})$ by a new pair $(Y^{\prime}_j, X^{\prime}_j)$ satisfying $(Y^{\prime}_j,X^{\prime}_j) \buildrel d \over = ({Y_j},{X_j})$, i.e. in distribution. Then we claim that the value of $Z_n({\beta ^*})$ minus the replaced version ${Z'_n}({\beta ^*})$ is at most $4L/n{w_{\min }}$. To see this, let
$$\mathbb{P}_{n}=:\dfrac{1}{n}\sum_{i=1}^{n}1_{Y_{i},X_{i}} ~~\text{and}~~\mathbb{P}_{n}^{'}=:\dfrac{1}{n}(\sum_{i=1, i\neq j}^{n}1_{Y_{i},X_{i}}+1_{Y_{j}^{'},X_{j}^{'}}),$$
where the second measure is the empirical measure corresponding to the replaced pair $(Y^{\prime}_j, X^{\prime}_j)$.

Then we have
\begin{align*}
{Z_n}({\beta ^*}) - {Z'_n}({\beta ^*}) & \le \underset{\beta\in \mathbb{R}^p}{\sup}\frac{{({\mathbb{P}_n} - \mathbb{P})\left\{l({\beta ^*}) - l(\beta )\right\} - ({\mathbb{P}{'}_n} - \mathbb{P})\left\{l({\beta ^*}) - l(\beta )\right\}}}{{||W(\beta  - {\beta ^*})|{|_1} + \varepsilon }}\\
& =\underset{\beta\in \mathbb{R}^p}{\sup}\frac{1}{n} \cdot \frac{{\left\{l({Y_i},{\beta ^*},{X_i}) - l({Y_i}, \beta ,{X_i})\right\} - \left\{l({{Y'}_i},{\beta ^*},{{X'}_i}) - l({{Y'}_i}, \beta ,{{X'}_i})\right\}}}{{||W(\beta  - {\beta ^*})|{|_1} + \varepsilon }}\\
& =\underset{\beta\in \mathbb{R}^p}{\sup}\frac{1}{n} \cdot \frac{- {Y_i}X_i({\beta ^*} - \beta ) + Y^{\prime}_i X^{\prime}_i({\beta ^*} - \beta ) - \frac{{{e^{X_i{\tilde \beta } }}}}{{1 + {e^{X_i{\tilde \beta }}}}}  X_i({\beta ^*} - \beta ) + \frac{{{e^{X^{\prime}_i{\tilde \beta } }}}}{{1 + {e^{X^{\prime}_i{\tilde \beta } }}}} X^{\prime}_i({\beta ^*} - \beta )}{{||W(\beta  - {\beta ^*})||_1 + \varepsilon }}\\
& \le \underset{\beta\in \mathbb{R}^p}{\sup}\frac{1}{n} \cdot \frac{{4L||\beta  - {\beta ^*}|{|_1}}}{{{w_{\min }}||\beta  - {\beta ^*}|{|_1} + \varepsilon }}\le \frac{{4L}}{{n{w_{\min }}}}.
\end{align*}
where the first inequality stems from for any function $f,g$
\[\left| {f(x)} \right| - \mathop {\sup }\limits_x \left| {g(x)} \right| \le \left| {f(x) - g(x)} \right| \Rightarrow \mathop {\sup }\limits_x \left| {f(x)} \right| - \mathop {\sup }\limits_x \left| {g(x)} \right| \le \mathop {\sup }\limits_x \left| {f(x) - g(x)} \right|.\]

Thus we could employ the McDiarmid's inequality in Lemma~\ref{lm:bd} by the achieving bounded difference condition. Therefore, we obtain
\begin{equation}\label{eq:McDiarmid}
P\left[{Z_n}({\beta ^*}) \ge x + {\rm{E}}\left\{{Z_n}({\beta ^*})\right\}\right]= P\left[{Z_n}({\beta ^*}) - {\rm{E}}\left\{{Z_n}({\beta ^*})\right\} \ge x\right]\le {\rm{exp}}\left( - \frac{{n{x^2}{w_{{{\min }}}^2}}}{{8{L^2}}}\right).
\end{equation}

It sufficient to get the upper bound of ${\rm{E}}\left\{{Z_n}({\beta ^*})\right\}$,
\begin{equation}\label{eq:EZ}
{\rm{E}}\left\{{Z_n}({\beta ^*})\right\} \le \frac{{8LA}}{{{w_{\min }}}}\sqrt {\frac{{2\log (2p)}}{n}},~~(A\ge1) .
\end{equation}

To obtain (\ref{eq:EZ}), we need the following two lemmas. The proof of the following symmetrization and contraction theorem can be found in Section 14.7 of \cite{Buhlmann2011}.

Let $\textit{X}_{1},...,\textit{X}_{n}$ be independent random variables taking values in some space $\mathcal{X}$ and $\mathcal{F}$ be a class of real-valued functions on $\mathcal{X}$.

\begin{lemma}[Symmetrization Theorem]\label{lm:Symmetrization}
Let $\varepsilon_{1},...,\varepsilon_{n}$ be a Rademacher sequence with uniform distribution on $\{ - 1,1\}$, independent of $\textit{X}_{1},...,\textit{X}_{n}$ and $f\in \mathcal{F}$. Then we have
$${\rm{E}}\left[ \underset{f \in \mathcal{F}}{\sup}\left\lvert \sum_{i=1}^{n}\left[ f(\textit{X}_{i})-{\rm{E}}\left\{f(\textit{X}_{i})\right\} \right] \right\rvert\right]\le 2{\rm{E}}\left[{\rm{E}}_{\epsilon}\left\{ \underset{f \in \mathcal{F}}{\sup}\left\lvert \sum_{i=1}^{n} \epsilon_{i}f(\textit{X}_{i}) \right\rvert\right\}\right]. $$
where ${\rm{E}}[\cdot]$ refers to the expectation w.r.t. $\textit{X}_{1},...,\textit{X}_{n}$ and ${\rm{E}}_{\epsilon}\left\{\cdot\right\}$ w.r.t. $\epsilon_{1},...,\epsilon_{n}$.
\end{lemma}

\begin{lemma}[Contraction Theorem]\label{lm:Contraction}
Let $x_{1},...,x_{n}$ be the non-random elements of $\mathcal{X}$ and $\varepsilon_{1},...,\varepsilon_{n}$ be Rademacher sequence. Consider $c$-Lipschitz functions $g_{i}$, i.e. $\left| {{g_i}(s) - {g_i}(t)} \right| \le c \left| {s - t} \right|,\forall s,t \in R$. Then for any function $f$ and $h$ in $\mathcal{F}$, we have
\[{\rm{E}}_{\epsilon}\left[\mathop {\sup }\limits_{f \in {\mathcal F}} \left\lvert {\sum\limits_{i = 1}^n {{\varepsilon_i}} \left[ {{g_i}\left\{f({x_i})\right\} - {g_i}\left\{h({x_i})\right\}} \right]} \right\rvert{\rm{ }}\right] \le 2c {\rm{E}}_{\epsilon}\left[\mathop {\sup }\limits_{f \in {\mathcal F}} \left\lvert {\sum\limits_{i = 1}^n {{\varepsilon_i}\left\{f({x_i}) - h({x_i})\right\}} } \right\rvert\right].\]
\end{lemma}

Note that ${({\mathbb{P}_n} - \mathbb{P})\left\{{l}({\beta ^*}) - {l}( \beta )\right\}}={{\mathbb{P}_n}\left\{{l}({\beta ^*}) - {l}( \beta )\right\}}-{\rm{E}}{\left\{{l}({\beta ^*}) - {l}( \beta )\right\}}$.  Next, for
$$nZ_{n}({\beta ^*}) = \mathop {\sup }\limits_{\boldsymbol{ \beta} \in \mathbb{R}^p} \left\{ \frac{{|\sum\limits_{k = i}^n [(  - Y{X}{\beta ^*} + \log \left\{1 + e^{X{\beta ^*}} \right\})-(  - Y{X}\beta  + \log \left\{1 + e^{X\beta} \right\})]-n{\rm{E}}{[{l_2}({\beta ^*}) - {l_2}( \beta )]}|}}{{||W(\beta  - {\beta ^*})||_1 + {\varepsilon _n}}}\right\}$$
as the suprema of the normalized empirical process (a local random Lipschitz constant), it is required to check the Lipschitz property of $g_i$ in Lemma~\ref{lm:Contraction} with ${\mathcal F}=\mathbb{R}^p$. Let $f({x_i})= \frac{{x_i^T} {\beta }}{{{||W(\beta  - {\beta ^*})||_1 + {\varepsilon _n}}}}$ and ${g_i}(t) = \frac{{{Y_i}(t{{||W(\beta  - {\beta ^*})||_1 + t{\varepsilon _n}}}) - \log (1 + {e^{t{{||W(\beta  - {\beta ^*})||_1 + t{\varepsilon _n}}}}})}}{{||W(\beta  - {\beta ^*})||_1 + {\varepsilon _n}}}$ with $t\in [-LB,LB]$. Then
\[g_i(s)-g_i(t) ={{ - {Y_i}(s - t) + \frac{{{e^{\tilde t }}}}{{1 + {e^{\tilde t }}}}(s - t)}} \le 2{|s - t|},~t,s\in [-LB,LB]\]
where $\tilde t$ is an intermediate point between $s$ and $t$ given by applying Lagrange mean value theorem for function ${g_i}(t)$.

Thus the function $g_i$ here is $2$-Lipschitz (in the sense of Lemma~\ref{lm:Contraction}).

Apply the symmetrization theorem and the contraction theorem, it implies
\begin{align*}
{\rm{E}}\left\{{Z_{n}({\beta ^*})}\right\} & \le  \frac{8}{n} {\rm{E}} \left\{\underset{\beta\in \mathbb{R}^p}{\sup}\sum_{i=1}^{n}\left\lvert \varepsilon_{i} \frac{{\textit{X}_{i}}^{T}(\beta^{*}-\beta)}{{\|{W(\beta  - {\beta ^*})}\|_1 + {\varepsilon _n}}}\right\rvert \right\}\\
[\text{Due to H{\"o}lder's inequality}]~&  \le \frac{8}{n} {\rm{E}} \left(\underset{\beta\in \mathbb{R}^p}{\sup}\mathop {\max }\limits_{1 \le j \le p}\left\lvert\sum_{i=1}^{n}  {\epsilon_{i}\textit{X}_{ij}}  \right\rvert \cdot \frac{\|{\beta  - {\beta ^*}}\|_1}{\|{W(\beta  - {\beta ^*})}\|_1 + {\varepsilon _n}}\right) \\
& =\frac{8}{{n{w_{\min }}}}{\rm{E}}\left({\rm{E}}_{\epsilon} \mathop {\max }\limits_{1 \le j \le p}\left\lvert\sum_{i=1}^{n}  {\epsilon_{i}\textit{X}_{ij}} \right\rvert \right).
\end{align*}

By using the maximal inequality (for examples, see Proposition A.1 in \cite{Zhang2017}), with ${\rm{E}}[{\epsilon_{i}\textit{X}_{ij}}|X]=0$ for all $i$, we get
\begin{align*}
{\rm{E}}\left\{{Z_{n}({\beta ^*})}\right\} \le\frac{8}{{n{w_{\min }}}} {\rm{E}}\left({\rm{E}}_{\epsilon} \mathop {\max }\limits_{1 \le j \le p}\left\lvert\sum_{i=1}^{n}  {\epsilon_{i}\textit{X}_{ij}} \right\rvert \right)&\le \frac{8}{{n{w_{\min }}}}\sqrt {2\log 2p} {\rm{E}}( \sqrt {nL^2})\le \frac{{8LA}}{{{w_{\min }}}}\sqrt {\frac{{2\log (2p)}}{n}} ,
\end{align*}
for $A\ge 1$. Thus, we obtain the upper bound of ${\rm{E}}\left\{{Z_{n}({\beta ^*})}\right\}$ as equation (\ref{eq:EZ}).

In equation (\ref{eq:McDiarmid}), if we choose $x$ such that ${\rm{exp}}\left\{-nx^2w_{\min}^2/\left(8L^2\right)\right\}=(2p)^{-A^2}$, then we get $x=\frac{{2LA}}{{{w_{\min }}}}\sqrt {\frac{{2\log (2p)}}{n}} $. By equation (\ref{eq:McDiarmid}) and equation (\ref{eq:EZ}), it implies
\[P\left\{Z_{n}({\beta ^*})\ge \frac{{2LA}}{{{w_{\min }}}}\sqrt {\frac{{2\log (2p)}}{n}}  + \frac{{8LA}}{{{w_{\min }}}}\sqrt {\frac{{2\log (2p)}}{n}} \right\}\le P\left[{Z_n}({\beta ^*}) \ge x + {\rm{E}}\left\{{Z_n}({\beta ^*})\right\}\right] \le {(2p)^{ - {A^2}}}.\]
Since $\hat \beta \in \mathbb{R}^p$, the last inequality gives by the definition of $Z_{n}({\beta ^*})$
\[P\left\{\frac{{({\mathbb{P}_n} - \mathbb{P})\{ {{l}({\beta ^*}) - {l}(\hat\beta)} \}}}{{ ||W( \hat\beta  - {\beta ^*})|{|_1}  + {\varepsilon }}}\ge \frac{{10LA}}{{{w_{\min }}}}\sqrt {\frac{{2\log (2p)}}{n}} \right\} \le P\left\{Z_{n}({\beta ^*})\ge \frac{{10LA}}{{{w_{\min }}}}\sqrt {\frac{{2\log (2p)}}{n}} \right\} \le {(2p)^{ - {A^2}}}.\]
Let $\frac{\lambda }{2}  \ge \frac{{10LA}}{{{w_{\min }}}}\sqrt {\frac{{2\log (2p)}}{n}} $, we have
\begin{equation}\label{eq:lambda}
({\mathbb{P}_n} - \mathbb{P})\{{l}({\beta ^*}) - {l}(\hat \beta )\} \le \frac{\lambda }{2} \{||W(\hat \beta  - {\beta ^*})|{|_1} + {\varepsilon _n}\}
\end{equation}
holding with probability at least $1 - {(2p)^{ - {A^2}}}$.

\subsection{Step2: Check $\hat{\beta}-\beta^{*} \in WC(3,\varepsilon_{n})$}

On the event (\ref{eq:lambda}), we can turn the inequality (\ref{eq:def-putting}) into
\begin{equation}\label{eq:lambda1}
\begin{aligned}
 \mathbb{P}\{l(\hat \beta )- l(\beta^{*})\}+ \frac{\lambda}{2}  ||W(\hat \beta  - {\beta ^*})||_1 & \le ({\mathbb{P}_n} - \mathbb{P})\{l({\beta ^*})-l(\hat \beta )\} +\frac{\lambda}{2}  ||W(\hat \beta  - {\beta ^*})||_1+ \lambda (||W{\beta ^*} |{|_1} - ||W\hat \beta|{|_1})\\
& \le \frac{\lambda }{2}\{||W(\hat \beta  - {\beta ^*})|{|_1} + {\varepsilon _n}\}+ \frac{\lambda}{2}  ||W(\hat \beta  - {\beta ^*})||_1+ \lambda (||W{\beta ^*} |{|_1} - ||W\hat \beta|{|_1}).
\end{aligned}
\end{equation}

By the definition of $\beta^*$, we have $\mathbb{P}\{l(\hat \beta )- l(\beta^{*})\}>0$, the above inequality reduces to
\begin{align}
\frac{\lambda }{2} ||W(\hat \beta  - {\beta ^*})||_1 & \le \frac{\lambda }{2}{\varepsilon _n}+ \lambda ||W(\hat \beta  - {\beta ^*})||_1+ \lambda (||W{\beta ^*} |{|_1} - ||W\hat \beta|{|_1})\label{eq-WC01}\\
& \le \frac{\lambda }{2}{\varepsilon _n}+ 2 \lambda ||W_{H}(\hat \beta  - {\beta ^*})_{H}||_1,\label{eq-WC1}
\end{align}
where $W_{H}$ is a diagonal matrix with diagonal element $w_{jj}=0$ if $j\notin H$ and $w_{jj}$ if $j\in H$; the last inequality is obtained by the fact that $|w_j{{\hat \beta }_j} - w_j \beta _j^*|{\rm{ + }}|w_j\beta _j^*| - |w_j{{\hat \beta }_j}|{\rm{ = 0}}$ for $j\notin H$ and $|w_j{{\hat \beta }_j}| - |w_j\beta _j^*| \le |w_j{{\hat \beta }_j} - w_j \beta _j^*|$ for $j\in H$.

From equation (\ref{eq-WC1}), we get
\begin{equation}\label{eq-WC}
\frac{\lambda }{2} ||W_{H^c}(\hat \beta  - {\beta ^*})_{H^c}||_1 \le \frac{3\lambda}{2} ||W_{H}(\hat \beta  - {\beta ^*})_{H}||_1 + \frac{\lambda }{2}{\varepsilon _n}.
\end{equation}

Therefore, we conclude that $\hat{\beta}-\beta^{*} \in WC(3,\varepsilon_{n})$.

\subsection{Step3: Derive error bounds from Stabil Condition}
The equation \eqref{eq-WC01} implies
\begin{equation}\label{eq:lambda2}
\begin{aligned}
\lambda {w_{\min }} ||\hat \beta  - {\beta ^*}||_1 \le \lambda ||W(\hat \beta  - {\beta ^*})||_1 & \le 2\lambda||W(\hat \beta  - {\beta ^*})|{|_1} + {\lambda {\varepsilon _n}}+  2\lambda (||W{\beta ^*} |{|_1} - ||W\hat \beta|{|_1})\\
& \le 2\lambda ||W\hat \beta ||_1 + 2\lambda ||W{\beta ^*}||_1 + {\lambda {\varepsilon _n}}+  2\lambda (||W{\beta ^*} |{|_1} - ||W\hat \beta|{|_1})\\
& = {\rm{4}}\lambda ||W{\beta ^*}|{|_1} + {{\lambda {\varepsilon _n}}}\le {\rm{4}}\lambda {w_{\max }}||{\beta ^*}|{|_1} + {{\lambda {\varepsilon _n}}}.
\end{aligned}
\end{equation}

Therefore, by Assumption $2$, we have
\begin{equation}
||\hat \beta  - {\beta ^*}|{|_1} \le \frac{{{\rm{4}}{w_{\max }}}}{{{w_{\min }}}}||{\beta ^*}|{|_1} + \frac{{{\varepsilon _n}}}{w_{\min}} \le \frac{{{\rm{4}}B{w_{\max }}}}{{{w_{\min}}}} + \frac{{{\varepsilon _n}}}{w_{\min}}.\nonumber
\end{equation}

Let $X^*\tilde{\beta}$ be an intermediate point between $X^*\hat{\beta}$ and $X^*{\beta ^*}$ given by applying Lagrange mean value theorem for $f(t) = t/\left(1+e^t\right)$. From Assumption $1$ and Assumption $2$, we have
\begin{equation}\label{eq:bd}
|{X^*}\tilde \beta | \le |{X^*}\tilde \beta  - {X^*}{{\beta }^*}| + |{X^*}{{\beta }^*}| \le |{X^*}\hat \beta  - {X^*}{{\beta }^*}| + |{X^*}{{\beta }^*}| \le L\left\{\frac{{B\left(4{w_{\max }} + {w_{\min }}\right)}}{{{w_{\min }}}} + \frac{{{\varepsilon _n}}}{w_{\min}}\right\}.
\end{equation}

By equation \eqref{eq:bd} and condition \eqref{eq:star}, we have $| {{X^*}\tilde \beta }| \le L{B^*}$. Let $X^*\tilde{\beta}$ be an intermediate point between $X^*{\beta}$ and $X^*{\beta ^*}$ given by applying Lagrange mean value theorem for $f(t) = \log(1+e^t)$, where  ${\beta }\in \Lambda$. Since ${X^*}$ is an independent copy of ${X}$, we have
\begin{align}\label{equation-Steinwart}
\mathbb{P}\{l(\hat \beta ) - l({\beta ^*})\}&= {{\rm{E}}^{\rm{*}}}[{\rm{E}}\{l(\beta ) - l({\beta ^*})|{X^*}\}]|_{\beta=\hat \beta}={{\rm{E}}^{\rm{*}}}\left\{{\rm{E}} { {[ - YX(\beta  - {\beta ^*}) + \log \frac{{1 + {e^{X\beta }}}}{{1 + {e^{X{\beta ^*}}}}}]|{X^*}} } \right\}|_{\beta=\hat \beta}\nonumber\\
&= {\left. {{{\rm{E}}^{\rm{*}}}\left\{ {{\rm{E}}[ - Y|{X^*}]{X^*}(\beta  - {\beta ^*}) + \log \frac{{1 + {e^{{X^*}\beta }}}}{{1 + {e^{{X^*}{\beta ^*}}}}}} \right\}} \right|_{\beta  = \hat \beta }}\nonumber\\
 ({{\rm{E}}^{\rm{*}}}[Y|{X^*}] = \frac{{{e^{{X^*}{\beta ^*}}}}}{{1 + {e^{{X^*}{\beta ^*}}}}})~ &= {{\rm{E}}^{\rm{*}}}{\left. {\left\{ {\frac{{ - {e^{{X^*}{\beta ^*}}}[{X^*}(\beta  - {\beta ^*})]}}{{1 + {e^{{X^*}{\beta ^*}}}}} + \frac{{{e^{{X^*}{\beta ^*}}}[{X^*}(\beta  - {\beta ^*})]}}{{1 + {e^{{X^*}{\beta ^*}}}}} + \frac{{{e^{{X^*}\tilde \beta }}{{[{X^*}(\beta  - {\beta ^*})]}^2}}}{{2{{(1 + {e^{{X^*}\tilde \beta }})}^2}}}} \right\}} \right|_{\beta  = \hat \beta }}\nonumber\\
&={{\rm{E}}^{\rm{*}}}{\left. {\left\{ {\frac{{{e^{{X^*}\tilde \beta }}}}{{2{{(1 + {e^{{X^*}\tilde \beta }})}^2}}}{{[{X^*}(\beta  - {\beta ^*})]}^2}} \right\}} \right|_{\beta  = \hat \beta }} \ge \mathop {\inf }\limits_{\left| t \right| \le LB} \frac{{{e^t}}}{{2{{(1 + {e^t})}^2}}}{{{\rm{E}}^{\rm{*}}}[{X^*}(\hat \beta  - {\beta ^*})]^2}\nonumber\\
& = \frac{{{e^{LB}}}}{{2{{(1 + {e^{LB}})}^2}}}{\rm{E^*}}[{{ X^{*}}( \hat \beta- {\beta ^*} )}]^2=: s{\rm{E^*}}[{{ X^{*}}( \hat \beta- {\beta ^*} )}]^2,
\end{align}
where $s = \frac{{{e^{LB}}}}{{2{{(1 + {e^{LB}})}^2}}}$.

Let $\Sigma= {\rm{E^*}}({ X^{*}}^T{ X^{*}})$ be the $p \times p$ covariance matrix. We have the expected prediction error:
\[{\rm{E^*}}{\{{ X^{*}}( \hat \beta- {\beta ^*} )\}^2} = {(\hat \beta  - {\beta ^*})}{\Sigma} (\hat \beta  - {\beta ^*}).\]

Since $WC(3,\varepsilon_{n})$ verifies the weighted cone condition, we thus could pose

1. Weighted Stabil Condition
\[s{(\hat \beta  - {\beta ^*})}{\Sigma}(\hat \beta  - {\beta ^*}) \ge sk||{W_H}{(\hat \beta  - {\beta ^*})_H}|{|_2^2} - s{\varepsilon _n}.\]

2. Stabil Condition
\[s{(\hat \beta  - {\beta ^*})^T}{\Sigma}(\hat \beta  - {\beta ^*}) \ge sk||{(\hat \beta  - {\beta ^*})_H}|{|_2^2} - s{\varepsilon _n}.\]

From equation (\ref{eq:lambda1}) and equation (\ref{eq-WC1}), we get
\begin{equation}\label{eq-WC2}
\mathbb{P}\{l(\hat \beta )- l(\beta^{*})\}+ \frac{\lambda }{2} ||W(\hat \beta  - {\beta ^*})||_1 \le \frac{\lambda }{2}{\varepsilon _n}+ 2 \lambda ||W_{H}(\hat \beta  - {\beta ^*})_{H}||_1 .
\end{equation}

\subsubsection{Case of Weighted Stabil Condition}
By equation (\ref{eq-WC2}) and the lower bound (\ref{equation-Steinwart}), we have
\begin{equation}\label{equation-CS0}
s{\rm{E^*}}[{\{{ X^{*}}( \hat \beta- {\beta ^*} )\}^2}]+ \frac{\lambda }{2} ||W(\hat \beta  - {\beta ^*})||_1 \le \frac{\lambda }{2}{\varepsilon _n}+ 2 \lambda ||W_{H}(\hat \beta  - {\beta ^*})_{H}||_1 .
\end{equation}
Substitute the inequality of the Weighted Stabil Condition into equation (\ref{equation-CS0}), we have
\begin{equation}
sk||{W_H}{(\hat \beta  - {\beta ^*})_H}|{|_2^2}+ \frac{\lambda }{2} ||W(\hat \beta  - {\beta ^*})||_1 \le \left(\frac{\lambda }{2}+s\right){\varepsilon _n}+ 2 \lambda ||W_{H}(\hat \beta  - {\beta ^*})_{H}||_1 .\nonumber
\end{equation}
Employing the Cauchy-Schwarz inequality, we have
\begin{equation}\label{equation-CS}
2sk||{W_H}{(\hat \beta  - {\beta ^*})_H}|{|_2^2} + \lambda ||W(\hat \beta  - {\beta ^*})|{|_1} \le (\lambda  + 2s){\varepsilon _n} + 4\lambda \sqrt {{d^*} ||{W_H}{{(\hat \beta  - {\beta ^*})}_H}||_2^2} .
\end{equation}
By applying the  elementary inequality $2xy \le tx^{2}+y^{2}/t$ to equation (\ref{equation-CS}) for all $t>0$, we have
\begin{equation}\label{equation-insertT}
2sk||{W_S}{(\hat \beta  - {\beta ^*})_S}|{|_2^2} + \lambda ||W(\hat \beta  - {\beta ^*})|{|_1} \le (\lambda  + 2s){\varepsilon _n} + 4t{\lambda ^2}{d^*} + {\textstyle{1 \over t}}||{W_H}{(\hat \beta  - {\beta ^*})_H}||_2^2.
\end{equation}
Let $t = {(2sk)^{ - 1}}$ in equation (\ref{equation-insertT}), thus
\begin{equation}
||W(\hat \beta  - {\beta ^*})|{|_1} \le 4t\lambda {d^*} + \left(\frac{{\lambda  + 2s}}{\lambda }\right){\varepsilon _n} = \frac{{2\lambda {d^*}}}{{sk}} + \left(\frac{{\lambda  + 2s}}{\lambda }\right){\varepsilon _n}.\nonumber
\end{equation}
Then
\begin{equation}
||\hat \beta  - {\beta ^*}|{|_1} \le \frac{{4\lambda {d^*}}}{{2sk{w_{\min }}}} + \left(\frac{{\lambda  + 2s}}{{\lambda {w_{\min }}}}\right){\varepsilon _n}.
\end{equation}

In order to derive the oracle inequality of $\ell_{2}$-prediction error, from (\ref{equation-CS0}) we have
\begin{align*}\label{eq:predict-error}
s{\rm{E^*}}[{\{{ X^{*}}( \hat \beta- {\beta ^*} )\}^2}] + \frac{\lambda }{2}||W(\hat \beta  - {\beta ^*})|{|_1} & \le \frac{\lambda }{2}{\varepsilon _n} + 2\lambda \{||W(\hat \beta  - {\beta ^*})|{|_1} - ||{W_{{H^c}}}{(\hat \beta  - {\beta ^*})_{{H^c}}}|{|_1}\}\\
& \le \frac{\lambda }{2}{\varepsilon _n} + 2\lambda ||W(\hat \beta  - {\beta ^*})|{|_1}.
\end{align*}
Then
\begin{equation}\label{equation-sE}
s{\rm{E^*}}[{\{{ X^{*}}( \hat \beta- {\beta ^*} )\}^2}] \le \frac{\lambda }{2}{\varepsilon _n} + \frac{3\lambda}{2} ||W(\hat \beta  - {\beta ^*})|{|_1}.
\end{equation}
Using the previous obtained $\ell_{1}$-estimation error bound for $||W(\hat \beta  - {\beta ^*})|{|_1}$, we have
\[s{\rm{E^*}}[{\{{ X^{*}}( \hat \beta- {\beta ^*} )\}^2}]\le \frac{\lambda }{2}{\varepsilon _n} + \frac{{3{\lambda ^2}{d^*}}}{{sk}} + \left(\frac{3\lambda}{2}  + 3s\right){\varepsilon _n} = \frac{{3{\lambda ^2}{d^*}}}{{sk}} + (2\lambda  + 3s){\varepsilon _n}.\]
Then we have
${\rm{E^*}}[{\{{ X^{*}}( \hat \beta- {\beta ^*} )\}^2}] \le \frac{{3{\lambda ^2}{d^*}}}{{{s^2}k}} + \left(\frac{{2\lambda }}{s} + 3\right){\varepsilon _n}.$

\subsubsection{Case of Stabil Condition}
By the inequality of Stabil Condition, equation (\ref{equation-CS0}) becomes
$$
sk||{(\hat \beta  - {\beta ^*})_H}|{|_2^2} + \frac{\lambda }{2}||W(\hat \beta  - {\beta ^*})|{|_1} \le \left(\frac{\lambda }{2} + s\right){\varepsilon _n} + 2\lambda ||{W_H}{(\hat \beta  - {\beta ^*})_H}|{|_1}.
$$
Employing the Cauchy-Schwarz inequality, we have
\begin{equation}\label{equation-1CS}
2sk||{(\hat \beta  - {\beta ^*})_H}||_2^2 + \lambda ||W(\hat \beta  - {\beta ^*})|{|_1} \le (\lambda  + 2s){\varepsilon _n} + 4\lambda \sqrt {||{W_H}|{|_2^2} \cdot ||{{(\hat \beta  - {\beta ^*})}_H}||_2^2} .
\end{equation}
By the elementary inequality $2xy \le tx^{2}+y^{2}/t$ to equation (\ref{equation-1CS}) for all $t>0$, then it derives
\begin{equation}\label{equation-insertT1}
2sk||{(\hat \beta  - {\beta ^*})_H}||_2^2 + \lambda ||W(\hat \beta  - {\beta ^*})|{|_1} \le (\lambda  + 2s){\varepsilon _n} + 4t{\lambda ^2}||{W_H}||_2^2 + \frac{1}{t}||{(\hat \beta  - {\beta ^*})_H}||_2^2.
\end{equation}
Let $t = {(2sk)^{ - 1}}$ in equation \eqref{equation-insertT1}, thus
\begin{equation}
||W(\hat \beta  - {\beta ^*})|{|_1} \le 4t\lambda ||{W_H}||_2^2 + \left(\frac{{\lambda  + 2s}}{\lambda }\right){\varepsilon _n} = \frac{{2\lambda ||{W_H}||_2^2}}{{sk}} + \left(\frac{{\lambda  + 2s}}{\lambda }\right){\varepsilon _n}.\nonumber
\end{equation}
Consequently,
\[||\hat \beta  - {\beta ^*}|{|_1} \le \frac{{2\lambda ||{W_H}||_2^2}}{{sk{w_{\min }}}} + \left(\frac{{\lambda  + 2s}}{{\lambda {w_{\min }}}}\right){\varepsilon _n}.\]

To obtain the oracle inequality of $\ell_{2}$-prediction error, it derives by equation (\ref{equation-sE}) and the previous obtained $\ell_{1}$-estimation error bound for $||W(\hat \beta  - {\beta ^*})|{|_1}$ that
\[s{\rm{E}}{\{{X^T}({\beta ^*} - \hat \beta )\}^2} \le \frac{\lambda }{2}{\varepsilon _n} + \frac{{3{\lambda ^2}||{W_H}||_2^2}}{{sk}} + \left(\frac{3\lambda}{2}  + 3s\right){\varepsilon _n} = \frac{{3{\lambda ^2}||{W_H}||_2^2}}{{sk}} + (2\lambda  + 3s){\varepsilon _n}.\]
Thus we get
${\rm{E}}{\{{X^T}({\beta ^*} - \hat \beta )\}^2} \le \frac{{3{\lambda ^2}{||{W_H}||_2^2}}}{{{s^2}k}} + \left(\frac{{2\lambda }}{s} + 3\right){\varepsilon _n}.$

\subsection{Proof of Corollary \ref{cl:ci}}
\begin{proof}
It is directly followed from the inequality
\begin{align*}
P(H \subsetneqq \hat H) &\le P(j \in \hat{H} \text { for some } j \in H)\\
&\le P(\hat{\beta}_{j}=  0 \text { and } \beta_{j}^{*}\neq 0, \text { for some } j \in H)\\
&\le P(|\tilde{\beta}_{j}-\beta_{j}^{*}|= |\beta_{j}^{*}|, \text { for some } j \in H)\\
&\le P\left(\|\hat \beta  - {\beta ^*}\|_1 \geq \mathop {\min }\limits_{j \in H} |\beta _j^*|\right) \\
& \le P\left(\|\hat \beta  - {\beta ^*}\|_1 \geq {B_0}\right) \le {(2p)^{ - {A^2}}}= \delta.
\end{align*}
Solve the equation ${(2p)^{ - {A^2}}} = \delta $ for $p$, we get $p = \frac{1}{2}\exp \{ \frac{1}{{{A^2}}}\log \frac{1}{\delta }\} $.
\end{proof}

\section{Summary}
We compare several weighted Lasso methods to estimate sparse parameter vector in high-dimensional logistic regression, which lies in the data-dependent estimators $\hat\beta=:\hat\beta(W)$ based on the given weighted vector $W=:(w_1,\cdots,w_p)^T$. The number of covariates $p$ can be very large, even larger than the sample size $n$. We assume that the regression coefficient vector $\beta^{*}$ is sparse, that is, there are very few coordinate components of $\beta^{*}$ that are non-zero, and their corresponding index set is $S(\beta^{*}) \in\{1,\cdots,p\}$. By analyzing the KKT conditions and applying bounded difference concentration inequalities, we could have an optimal and simple data-adaptive weights as the number of covariates $p \to \infty $. Based on the Weighted Stabil Condition or Stabil Condition, we give two versions of non-asymptotic oracle inequalities for the weighted Lasso estimator in light of KKT conditions which are both guaranteed with high probability as $p \to \infty$. Our goal is to estimate the true vector $\beta^{*}$ by applying the weighted $\ell_1$-penalized estimator $\hat\beta$ to this unknown sparsity of $\beta^{*}$ and then show the probability of correct subset selection is high.

\section{Acknowledgements}
Three authors Huamei Huang, Yujing Gao and Huiming Zhang are co-first authors which contributes equally to this work. The authors would like to thank the editor, the associate editor, and the anonymous referees, whose insightful comments and constructive suggestions have greatly improved an earlier version of this manuscript.

%The  author of this paper  would like to thank  Acta Mathematica Sinica  for their help during writting  actams.cls


\begin{thebibliography}{99}

\bibitem[Algamal and Lee(2017)]{Algamal2017}
Algamal, Z. Y., Lee, M. H. (2017). A new adaptive L1-norm for optimal descriptor selection of high-dimensional QSAR classification model for anti-hepatitis C virus activity of thiourea derivatives. SAR and QSAR in Environmental Research, 28(1), 75-90.

\bibitem[{Bickel et al.(2009)}]{Bickel09}
Bickel, P. J., Ritov, Y. A., Tsybakov, A. B. (2009). Simultaneous analysis of Lasso and Dantzig selector. The Annals of Statistics, 1705-1732.


\bibitem[Buhlmann and van de Geer(2011)]{Buhlmann2011}
Buhlmann, P., van de Geer, S. (2011). Statistics for high-dimensional data: methods, theory and applications. Springer.

\bibitem[{Boucheron et al.(2013)}]{Boucheron13}
Boucheron, S., Lugosi, G., Massart, P. (2013). Concentration inequalities: A nonasymptotic theory of independence. Oxford university press.

\bibitem[{Bunea(2008)}]{Bunea08}
Bunea, F. (2008). Honest variable selection in linear and logistic regression models via l1 and l1+ l2 penalization. Electronic Journal of Statistics, 2, 1153-1194.

\bibitem[{Cox(1958)}]{Cox58}
Cox, D. R. (1958). The regression analysis of binary sequences (with discussion). Journal of the Royal Statistical Society. Series B (Methodological), 215-242.

\bibitem[{Dudoit et. al(2002)}]{Dudoit2002}
Dudoit, S., Fridlyand, J., \& Speed, T. P. (2002). Comparison of discrimination methods for the classification of tumors using gene expression data. Journal of the American statistical association, 97(457), 77-87.

\bibitem[Efron and Hastie(2016)]{Efron2016}
Efron, B., Hastie, T. (2016). Computer age statistical inference: algorithms, evidence, and data science. Cambridge University Press.

\bibitem[{Fan et al.(2020)}]{Fan20}
Fan, Y., Zhang, H., \& Yan, T. (2020). Asymptotic theory for differentially private generalized ¦Â-models with parameters increasing. Statistics and Its Interface, 13(3), 385-398.

\bibitem[{Golub et al.(1999)}]{Golub1999}
Golub, T. R., Slonim, D. K., Tamayo, P., Huard, C., Gaasenbeek, M., Mesirov, J. P., Caligiuri, M. A. (1999). Molecular classification of cancer: class discovery and class prediction by gene expression monitoring. Science, 286(5439), 531-537.

\bibitem[{Guo et al.(2015)}]{Guo15}
Guo, P., Zeng, F., Hu, X., Zhang, D., Zhu, S., Deng, Y., \& Hao, Y. (2015). Improved variable selection algorithm using a LASSO-type penalty, with an application to assessing Hepatitis B infection relevant factors in community residents. PloS one, 10(7), e0134151.

\bibitem[{Hastie et al.(2015)}]{Hastie15}
Hastie, T., Tibshirani, R., Wainwright, M. (2015). Statistical learning with sparsity: the lasso and generalizations. CRC Press.

\bibitem[Li and Lederer(2019)]{Li2019}
Li, W., \& Lederer, J. (2019). Tuning parameter calibration for l1-regularized logistic regression. Journal of Statistical Planning and Inference, 202, 80-98.

\bibitem[Liu and San Wong(2019)]{Liu2017}
Liu, C., \& Wong, H. S. (2019). Structured Penalized Logistic Regression for Gene Selection in Gene Expression Data Analysis. IEEE/ACM Transactions on Computational Biology and Bioinformatics, 16(1), 312-321.

\bibitem[{Kwemou(2016)}]{Kwemou16}
Kwemou, M. (2016). Non-asymptotic oracle inequalities for the Lasso and Group Lasso in high dimensional logistic model. ESAIM: Probability and Statistics, 20, 309-331.

\bibitem[Ma et al.(2020)]{Ma2018}
Ma, R., Tony Cai, T., \& Li, H. (2020). Global and Simultaneous Hypothesis Testing for High-Dimensional Logistic Regression Models. Journal of the American Statistical Association, 1-15.

\bibitem[Park and Konishi(2016)]{Park2016}
Park, H., Konishi, S. (2016). Robust logistic regression modelling via the elastic net-type regularization and tuning parameter selection. Journal of Statistical Computation and Simulation, 86(7), 1450-1461.

\bibitem[{Rigollet and H{\"u}tter(2019)}]{Rigollet19}
Rigollet, P., \& H{\"u}tter, J. C. (2019). High dimensional statistics. \url{http://www-math.mit.edu/~rigollet/PDFs/RigNotes17.pdf}

\bibitem[{Sur et al.(2019)}]{Sur17}
Sur, P., Chen, Y., \& Candes, E. J. (2019). The likelihood ratio test in high-dimensional logistic regression is asymptotically a rescaled Chi-square. Probability Theory and Related Fields, 175(1), 487¨C558.


\bibitem[{Tutz(2011)}]{Tutz11}
Tutz, G. (2011). Regression for categorical data. Cambridge University Press.

\bibitem[Tibshirani(1996)]{Tibshirani1996}
Tibshirani, R. (1996). Regression shrinkage and selection via the lasso. Journal of the Royal Statistical Society. Series B (Methodological), 267-288.

\bibitem[{van de Geer(2008)}]{Geer2008}
van de Geer, S. A. (2008). High-dimensional generalized linear models and the lasso. The Annals of Statistics, 614-645.

\bibitem[{Yang et al.(2019)}]{Yang19}
Yang, X., Zhang, H., Wei, H., \& Zhang, S. (2019). Sparse Density Estimation with Measurement Errors. arXiv preprint arXiv:1911.06215.

\bibitem[{Yin(2020)}]{Yin2020}
Yin, Z. (2020). Variable selection for sparse logistic regression. Metrika, 1-16.

\bibitem[{Zou(2006)}]{Zou2006}
Zou, H. (2006). The adaptive lasso and its oracle properties. Journal of the American statistical association, 101(476), 1418-1429.

\bibitem[{Zhang and Jia(2020)}]{Zhang2017}
Zhang, H., Jia, J. (2020). Elastic-net Regularized High-dimensional Negative Binomial Regression: Consistency and Weak Signals Detection. Statistica Sinica.

\bibitem[{Zhang(2018)}]{Zhang2018}
Zhang, H. (2018). A note on" MLE in logistic regression with a diverging dimension". arXiv preprint arXiv:1801.08898.

\end{thebibliography}
\end{document}